\pdfoutput=1
\documentclass[11pt]{article}
\usepackage{fullpage}
\usepackage[T1]{fontenc}
\usepackage{fourier}

\usepackage[cmex10]{amsmath}
\usepackage{bm}
\usepackage{amssymb}

\usepackage{algorithm}
\usepackage[noend]{algpseudocode}

\usepackage{tikz}
\usepackage{xcolor}
\usepackage{makeidx}
\usepackage{graphicx}
\usepackage{multirow}
\usepackage{hhline}
\usepackage{booktabs}       % professional-quality tables
\usepackage{amsmath}

\usepackage{wrapfig}
\usepackage{graphicx}
\usepackage{caption}
\usepackage{subcaption}
\graphicspath{ {data-plots/} }
\usepackage{mathtools}

\usepackage{url}

\newtheorem{example}{Example}
\newtheorem{definition}{Definition}

\makeatletter
\def\@BracContents{} % default (used by \suchthat)
\newcommand{\BracKern}{\kern-\nulldelimiterspace}
\newcommand{\@Brac}[3]{% #1,#3 = left/right bracket type
    \ensuremath{%
        \begingroup\def\@BracContents{#2}%
        \mathopen{\left#1\vphantom{#2}\BracKern\right.}% left bracket
        #2%  content
        \mathclose{\left.\BracKern\vphantom{#2}\right#3}% right bracket
        \endgroup%
    }%
}
\newcommand{\bracr}[1]{\@Brac{(}{#1}{)}}%
\makeatother

\newcommand{\ourModel}{\textsf{JoinInfer}}
\newcommand{\ourModelDown}{\textsf{JoinInferDown}}
\newcommand{\ourModelUp}{\textsf{JoinInferUp}}
\newcommand{\ourModelSampling}{\textsf{HYJAR}}

\newcommand{\NPRR}{\textsf{MultFacProd}}
\newcommand{\parent}{\textsf{Parent}}
\newcommand{\children}{\textsf{Children}}
\newcommand{\fhtw}{\mathrm{fhtw}}
\newcommand{\htw}{\mathrm{htw}}

\newenvironment{newEdits}{%
    \setlength{\parindent}{0pt}
}{}

\newcommand{\bandone}{Band $1$}
\newcommand{\bandtwo}{Band $2$}
\newcommand{\bandthree}{Band $3$}
\newcommand{\bandfour}{Band $4$}
\newcommand{\bandfive}{Band $5$}
\newcommand{\bandsix}{Band $6$}
\newcommand{\PP}{\textsf{PairwiseProd}}
\newcommand{\subtree}{\mathrm{subtree}}
\newcommand{\hash}{\mathrm{HashProduct}}
\newcommand{\random}{\mathrm{Random}}
\newcommand{\tw}{\mathrm{tw}}
\title{\textbf{Hypertree Decompositions Revisited for PGMs}}
\author{Aarthy Shivram Arun\footnotemark[1] \and Sai Vikneshwar Mani Jayaraman\footnotemark[1] \and Christopher R\'{e}\footnotemark[2] \and Atri Rudra\footnotemark[1]}
\date{\footnotemark[1]~ University at Buffalo, SUNY\\
\texttt{\{ashivram,saivikne,atri\}@buffalo.edu}\\
\vspace*{2mm}
\footnotemark[2]~ Stanford University\\
\texttt{chrismre@cs.standford.edu}
}

\begin{document}
%\onecolumn
%\maketitle
\maketitle
%\onecolumn
%\frenchspacing
%\setlength{\abovedisplayskip}{5pt}
%\setlength{\belowdisplayskip}{5pt}
\begin{abstract}
%We present JoinInfer, an exact inference engine that outperforms existing state-of-art inference engines (ACE, IJGP and libDAI) on PGMs defined on large domains and sparse factors (aka in the sense of determinism). Our engine implements recent theoretical improvements in exact inference algorithms, which are in turn based on recent \emph{worst-case optimal} database join algorithms. Unlike some existing engines that exploit determinism, we do not need a computationally expensive compilation phase. We also empirically demonstrate that properties of data that give improved theoretical results indeed determine JoinInfer's performance on UAI ad IJCAI datasets. Finally, we propose a simple sampling heuristic that allows JoinInfer to automatically tailor its parameters to the input data itself.

\begin{newEdits}
%\color{magenta}{
We revisit the classical problem of exact inference on probabilistic graphical models (PGMs). Our algorithm is based on recent \emph{worst-case optimal} database join algorithms, which
can be asymptotically faster than traditional data processing methods.
We present the first empirical evaluation of these new algorithms via {\ourModel}, a new exact inference engine. We empirically explore the properties of the data for which our engine can be expected to outperform traditional inference engines refining current theoretical notions. 
 Further, \ourModel\ outperforms existing state-of-the-art inference engines (ACE, IJGP and libDAI) on some standard benchmark datasets by up to a factor of 630x. Finally, we propose a promising data-driven heuristic that extends {\ourModel} to automatically tailor its
parameters and/or switch to the traditional inference algorithms. 
%This heuristic shows a lot of promise in attaining a `best of both worlds' inference architecture.
\end{newEdits}
%(ACE, IJGP and libDAI) 

%\commented out by asa{for inference problems. We
%construct a prototype, {\ourModel}, which is an exact inference engine
%that is able to outperforms existing state-of-art inference engines
%(ACE, IJGP and libDAI) on large domains and sparse factors by 3.5x on UAI and IJCAI datasets and at least 10x on synthetic datasets.}
%\yell{XX
%  PRECISE NUMBER XX}. 

%The theoretical advantages of these algorithms are especially pronounced in cyclical graphs: a bottleneck for exact methods. Further, unlike some existing engines that exploit determinism, we do not need a computationally expensive compilation phase. We also empirically demonstrate that properties of data that give improved theoretical results indeed determine JoinInfer's performance on UAI ad IJCAI datasets. Finally, we propose a simple sampling heuristic that allows JoinInfer to automatically tailor its parameters to the input data itself. 

\end{abstract}

\section{Introduction}
\begin{newEdits}

Efficient inference on probabilistic graphical models (PGMs) is a core topic in artificial intelligence (AI) and standard inference techniques are based on tree decomposition~\cite{jensen1990,dechter1996,kask2005,IJGP} . The runtime of such algorithms is exponential in the treewidth ($\tw$) of the underlying graph, which in the worst case, is unavoidable. Over the years, efforts in the logic, database and AI communities to refine $\tw$ into a finer-grained measure of complexity have culminated in {\em generalized hypertree decompositions} (GHDs)~\cite{gottlob2005,fischl2016}. Recently, FAQ/AJAR ~\cite{faq,ajar} theoretically reconnected such GHD-based algorithms with probabilistic inference and achieved tighter bounds based on a finer-grained notion of width called {\em fractional hyper-tree width} ($\fhtw$). %~\cite{jensen1990,shafer1990,dechter1996,kask2005,IJGP}

\paragraph{} However, (1) the practical significance of such GHD based PGM inference has met with some skepticism: Dechter et al.~\cite{dechter08}, via experimental evaluation, conclude that classical treewidth-based algorithms run faster on PGM benchmarks than those optimizing GHD-based measures. Their experiments suggest that the advantages of the latter manifest only in instances with substantial factor sparsity (i.e. large number of factor entries have zero probabilities) and high factor arity. (2) Translating the superior asymptotic bounds of GHDs into practice is a non-trivial challenge. In theory, these algorithms assume that one can exhaustively search over all potential GHDs, which is often untenable due to the combinatorial explosion of possible GHDs with thousands of variables and factors. Indeed, the theoretical runtimes of these algorithms completely ignore the dependence on the number of variables and factors; in practice, their asymptotic advantages may be negated by large constants. 

\paragraph{} In the current work, we revisit the conclusions in (1) and overcome the challenges of (2) using a proof-of-concept inference engine --- {\ourModel} --- that leverages recently introduced worst case optimal joins~\cite{ngo2} in conjunction with improved data structures. In particular, we make the following contributions:

\begin{figure}[t]
\begin{center}
\begin{tikzpicture}[scale = 1.9]

%Vertical Label
\node[above, rotate=90] at (.5,1.5) {{\small \textcolor{gray}{$\rho$ low}}};
\node[above, rotate=90] at (.5,2.5) {{\small \textcolor{gray}{$\rho$ high}}};

%Horizontal Lables
\node[above] at (1,3) {{\small \textcolor{gray}{$R_D$ small}}};
\node[above] at (2,3) {{\small \textcolor{gray}{$R_D$ medium}}};
\node[above] at (3,3) {{\small \textcolor{gray}{$R_D$ large}}};

%Legdent for R_D
\fill[color=red!90] (.5,.75) rectangle (.7,.95);
\node[right] at (.7,.85) {{\scriptsize \textcolor{gray}{$1$-$10^5$x \tiny fast}}};
\fill[color=red!60] (1.5,.75) rectangle (1.7,.95);
\node[right] at (1.7,.85) {{\scriptsize \textcolor{gray}{$1$-$10^{10}$x \tiny slow}}};
\fill[color=red!30] (2.5,.75) rectangle (2.7,.95);
\node[right] at (2.7,.85) {{\scriptsize \textcolor{gray}{$10^{10}$-$10^{20}$x \tiny slow}}};

%Legend for JI
\fill[color=blue!90] (.5,.5) rectangle (.7,.7);
\node[right] at (.7,.6) {{\scriptsize \textcolor{gray}{$10^2$-$10^3$x \tiny fast}}};
\fill[color=cyan!90] (2,.5) rectangle (2.2,.7);
\node[right] at (2.2,.6) {{\scriptsize \textcolor{gray}{$1$-$5$x \tiny fast}}};
\fill[color=cyan!40] (.5,.25) rectangle (.7,.45);
\node[right] at (.7,.36) {{\scriptsize \textcolor{gray}{$1$-$20$x \tiny (mean: 5x) slow}}};
\fill[color=cyan!10] (2,.25) rectangle (2.2,.45);
\node[right] at (2.2,.35) {{\scriptsize \textcolor{gray}{$1$-$20$x \tiny (mean: 10x) slow}}};

%Coloring of cells
%High rho
%band one
\fill[color=blue!90] (.5,2) rectangle (1,3);
\fill[color=red!90] (1,2) rectangle (1.5,3);
%band two
\fill[color=cyan!90] (1.5,2) rectangle (2,3);
\fill[color=red!60] (2,2) rectangle (2.5,3);
%band three
\fill[color=cyan!90] (2.5,2) rectangle (3,3);
\fill[color=red!30] (3,2) rectangle (3.5,3);
%Low rho
%band one
\fill[color=cyan!90] (.5,1) rectangle (1,2);
\fill[color=red!90] (1,1) rectangle (1.5,2);
%band two
\fill[color=cyan!40] (1.5,1) rectangle (2,2);
\fill[color=red!60] (2,1) rectangle (2.5,2);
%band three
\fill[color=cyan!20] (2.5,1) rectangle (3,2);
\fill[color=red!30] (3,1) rectangle (3.5,2);

%Lables for cells
%Band name
\node[right] at (.7,2.5) {{\footnotesize \textcolor{black}{\bandone}}};
%Winner engine
\node[right] at (.65,2.3) {{\scriptsize \textcolor{black}{(\ourModel)}}};
%Band name
\node[right] at (1.7,2.5) {{\footnotesize \textcolor{black}{\bandtwo}}};
%Winner engine
\node[right] at (1.65,2.3) {{\scriptsize \textcolor{black}{(\ourModel)}}};
%Band name
\node[right] at (2.7,2.5) {{\footnotesize \textcolor{black}{\bandthree}}};
%Winner engine
\node[right] at (2.65,2.3) {{\scriptsize \textcolor{black}{(\ourModel)}}};
%Band name
\node[right] at (.7,1.5) {{\footnotesize \textcolor{black}{\bandfour}}};
%Winner engine
\node[right] at (.65,1.3) {{\scriptsize \textcolor{black}{(\ourModel)}}};
%Band name
\node[right] at (1.7,1.5) {{\footnotesize \textcolor{black}{\bandfive}}};
%Winner engine
\node[right] at (1.72,1.3) {{\scriptsize \textcolor{black}{(libDAI)}}};
%Band name
\node[right] at (2.7,1.5) {{\footnotesize \textcolor{black}{\bandsix}}};
%Winner engine
\node[right] at (2.72,1.3) {{\scriptsize \textcolor{black}{(libDAI)}}};

%Draw the cells
%Low \rho boxes
%\draw[color=gray] (.5,1) -- (1.5,1) -- (1.5,2) -- (.5,2) -- (.5,1);
%\draw[color=gray] (1.5,1) -- (2.5,1) -- (2.5,2) -- (1.5,2) -- (1.5,1);
%\draw[color=gray] (2.5,1) -- (3.5,1) -- (3.5,2) -- (2.5,2) -- (2.5,1);
%Low \rho high
%\draw[color=gray] (.5,2) -- (1.5,2) -- (1.5,3) -- (.5,3) -- (.5,2);
%\draw[color=gray] (1.5,2) -- (2.5,2) -- (2.5,3) -- (1.5,3) -- (1.5,2);
%\draw[color=gray] (2.5,2) -- (3.5,2) -- (3.5,3) -- (2.5,3) -- (2.5,2);

\draw[color=gray] (.25,2) -- (3.5,2);
\draw[color=gray] (1.5,3.25) -- (1.5,1);
\draw[color=gray] (2.5,3.25) -- (2.5,1);

\end{tikzpicture}
\end{center}
\caption{\small{Datasets are divided  into six bands depending on the sizes of $R_D$ and $\rho$. The red shade in each box denotes the speedup of GHD based system over a treewidth based system and the blue shades shows the speedup of \ourModel\ with respect to libDAI. The ``winner" is stated explicitly for each band.}}
%\caption{\small{Runtime Scope of {\ourModel}: `Small' denotes `$10^{-6} < R_D \leq 1$', 'Medium' denotes `$1 < R_D \leq 10^{3}$', `Large' denotes `$10^{3} < R_D \leq 10^{20}$'.`JI' categorizes where {\ourModel} is the best engine and `LD' categorizes where libDAI is the best engine.} \yell{I have tokens on this figure-- Atri} \yell{Check: \bandone\ \bandtwo\ \bandthree\ \bandfour\ \bandfive\ \bandsix\ }}
\label{introTab}
\end{figure}
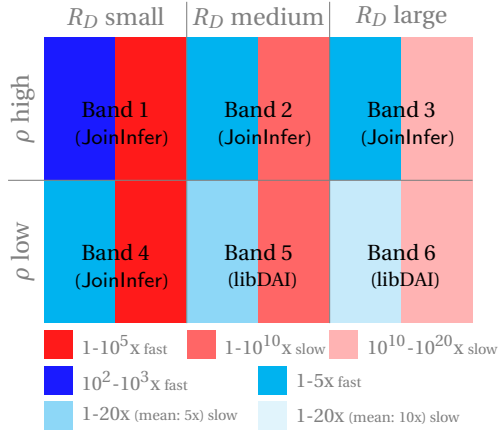

\paragraph{GHDs revisited.}

\begin{itemize}

\item Experimental evaluation in~\cite{dechter08} used a ratio of theoretical bounds based on the GHD measure of hypertree width ($\htw$) and the treewidth measure (we call the analogous version of this ratio $R_D$: where we replace $\htw$ with $\fhtw$).  However, we suggest that the predictions made by $R_D$ are, in practice, contingent upon the total number of entries processed across all bags of the GHD (we call this $\rho$). Engines such as libDAI that use truth-table indices do not scale to higher levels of $\rho$, an insight that is not captured in~\cite{dechter08}'s experimental paradigm. For instance, in the top row in Figure~\ref{introTab} (corresponding to high $\rho$),  libDAI fails on all datasets while \ourModel\ successfully completes on all of them.
%[cite Dechter]'s measure to evaluate the scope of GHD based algorithms was largely based on theoretical bounds 

%\item %A finer grained theoretical ratio (as compared to $R_D$) can be obtained using the sizes of individual factors (not $N$) and product of domain sizes of individual variables (not $D$) assigned to the bags. This new ratio better predicts {\ourModel}'s runtime.

%We obtain a finer grained theoretical ratio that is better at predicting {\ourModel}'s runtime: $R_J = \sum_B \prod_{e \in \mathcal{E}(B)} N_e ^{x_e}/\rho$, where $\mathcal{E}(B)$ are factors assigned to bag B, $N_e$ is their size and $0 \leq x_e\leq 1$ is a value that bounds the size of the factor product. %$R_J$ is better at predicting the runtimes of {\ourModel} as compared to $R_D$.

\item Introduction of $\rho$ shows that GHD based algorithms like {\ourModel} may have a wider scope than previously predicted: while~\cite{dechter08}  predicted that {\ourModel} will work well \emph{only} when $R_D$ is small, we expect it to perform well even when $R_D$ is medium-to-large (provided $\rho$ is high). We expect libDAI to perform well when $R_D$ is medium-to-large \emph{and} $\rho$ is low (Bands 5 and 6 in Figure~\ref{introTab}). 
 
\item We use a finer-grained theoretical measure that better predicts {\ourModel}'s speed up, thus refining the insights of~\cite{dechter08}. For instance, our measure can differentiate between the rows of column `$R_D$ small' in Figure~\ref{introTab}, while $R_D$ cannot.

%\item As Figure~\ref{introTab} shows, {\ourModel} can be upto 1000x faster on datasets where $R_D$ is small/medium and $\rho \geq 10^9$. Since, libDAI primarily relies on truth-table indexes which do not scale at these levels of $\rho$, the engine collapses. ACE is the only other engine that completes on a subset of the space since it is designed to accomodate higher levels of $\rho$. Further, {\ourModel} is upto $3.3x$ faster when $R_D$ is large and $\rho \geq 10^9$ and is 5x faster where $R_D$ is small and $\rho < 10^9$.

%\item We show that the scope of {\ourModel} is wider than previously predicted: it can be upto 1000x faster on networks characterized by a small to medium ratio measure $(R_D)$ and a high $\rho$. When $R_D$ is large (predicting extremely bad performance), at the higher threshold of $\rho$ {\ourModel} actually outperforms at non-trivial levels (by upto 2.7x). IJGP and libDAI fail in this space; ACE is the only other engine that completes on a subset of the space since it is designed to accomodate higher levels of $\rho$.

%\item We show that {\ourModel} can be upto 1000x faster on networks with small-to-medium $(R_D)$ and high $\rho$. When $R_D$ is large (which ~cite{dechter} predicts will correlate with underperformance by more than $10^3$x), at the higher threshold of $\rho$, {\ourModel} actually \emph{outperforms} at non-trivial levels (by upto 2.7x) when the prediction my. IJGP and libDAI fail in this space; ACE is the only other engine that completes on a subset of the space since it is designed to accomodate higher levels of $\rho$.

\item We show that {\ourModel} can be up to 630x faster on networks with small $R_D$ and high $\rho$ ({\bandone} in Figure~\ref{introTab}). When $R_D$ is large, at the higher threshold of $\rho$ ({\bandthree} in Figure~\ref{introTab}), {\ourModel} actually \emph{outperforms} at non-trivial levels (by upto 2.7x), while the corresponding prediction by~\cite{dechter08} expects under-performance by more than $10^{10}$x. IJGP and libDAI fail in this space; ACE is the only other engine that completes on a subset of the space, since it is designed to accommodate higher levels of $\rho$.

%(which ~cite{dechter} predicts will correlate with underperformance by more than $10^3$x), at the higher threshold of $\rho$, {\ourModel} actually \emph{outperforms} at non-trivial levels (by upto 2.7x) when the prediction my. IJGP and libDAI fail in this space; ACE is the only other engine that completes on a subset of the space since it is designed to accomodate higher levels of $\rho$.

%\item  libDAI's indexing and caching advantages manifest when $\rho < 10^9$ and $R_D$ is medium/large (the 2 settings which does not favor {\ourModel}). To counter this we introduce a hybrid architecture that uses a 'best-of-both-worlds' approach thus providing consistent performance even across these settings.

\end{itemize}

%(present better theoretical bounds
% claim by presenting a finer grained predictor of runtime complexity using the theoretical bounds introduced in FAQ/AJAR and show that these tighter theoretical bounds do translate to substantial runtime speed-ups in practice over a fairly large range of problem instances (that spread over a reasonable spectrum of determinism- do we need this?). %In some sense we show that sparsity is a sufficient condition.

%\item \textbf{Hybrid Architecture}
\paragraph{Hybrid Architecture.}

Given the relative advantages of {\ourModel} and libDAI in different spaces, we explore the feasibility of exploiting their strengths in a `best-of-all-worlds' architecture. The hybrid outperforms libDAI, IJGP and ACE for 75\% of the networks (ref. Table~\ref{table_real_3}), illustrating its promise.% in such fusions.
% libDAI's advantages manifest at lower levels of $\rho$ and when $R_D$ is medium to large (the 2 settings which do not favor {\ourModel}-- see the bottom middle and right cells in Figure~\ref{introTab}). To counter this, we introduce a hybrid architecture which adopts a 'best-of-both-worlds' approach thus providing consistent performance across multiple settings: always within x\% of the best engine.

%Notwithstanding the advantages of {\ourModel}, in %order to cover a wider array of problem instances: - (i)
 %large treewidth datasets, where the treewidth (\emph{tw}) bound is comparable to the fractional-hypertreewidth (\emph{fhtw}) bound, {\ourModel} does not perform as well as existing engines (in particular, libDAI).
% and (ii) networks with only binary valued variables with nearly "full" factor tables, i.e., absence of factor sparsity (advantageous to IJGP) - 
%We introduce a hybrid architecture that offers a `best-of-both worlds' approach. In this architecture we use data-driven heuristics to switch between existing engines (libDAI) and {\ourModel}, thus combining their respective advantages.

%\item \textbf{Technical Contributions}
\paragraph{Technical Contributions.}

%Theory says that this should be superior, but it is not that evident in practice...
%Although theory suggests that the recently introduced GHD based algorithms possess superior runtime bounds, whether these translate in practice is an open question. In theory, these algorithms assume that one can exhaustively search over all potential GHDs, which is untenable due to the combinatorial explosion of possible GHDs with thousands of variables and factors. Indeed, the theoretical runtimes of these algorithms completely ignore the dependence on the number of variables and factors-- in practice, their asymptotic advantages may be negated by large constants. Thus, we argue constructing a prototype and performing an empirical study is critical to understanding whether these algorithms can provide new ideas to this classical problem.

%We present {\ourModel}, an engine that leverages recently introduced worst case optimal joins (cite) in conjunction with improved data structures to perform exact inference. Our empirical results show that one can  improve the speed of exact inference on PGM datasets~\cite{uai2006,ijcai2005}, by up to 10x compared to state-of-the-art inference engines (libDAI, IJGP and ACE). 

%In  implementing {\ourModel}, we make the following technical contributions:

%Diverging from theory, to overcome the costs imposed by the conventional trie datastructure,  
Our primary technical contribution in {\ourModel} lies in improved data-\\structures. We introduce two data representations for use in different passes of the algorithm: (i) level-order trie, which collapses a conventional trie into a single array; (ii) (two variants of) an index-based compressed list. 
%These variants essentially store different indices, a step that enables us to re-use existing datastructures,  and,  optimize our algorithm. %simultaneously compute messages in the `down' pass. 
We find that the resulting gains more than compensate for the overheads involved in maintaining both data representations.

\end{newEdits}

\section{Related Work}
\label{sec:relWork}

%Efficient and accurate inference in probabilistic graphical models has been an actively researched area for over three decades, resulting in 
Several streams of inquiry have emerged in the exact inference setting. One such stream involves \emph{conditioning algorithms}~\cite{pearl1989,darwicheRecursive} that adopt a case-based reasoning approach. \emph{Cutset conditioning}~\cite{pearl1989} attempts to reduce a network into a tree structure so as to make inference tractable, while another approach \emph{recursive conditioning}~\cite{darwicheRecursive} recursively decomposes a network into smaller subnetworks that are solved independently.

%class of algorithms seek to exploit local structure such as determinism, context specific independences (CSI) etc., with a focus on improving model tractability [].
%Exploiting local structure has been a strategy adopted by several approaches [cite]
Another class of algorithms seeks to exploit local structure~\cite{LarkinDechter03,pooleZhang03}, where~\cite{chavira_07,huang06} exploit factor sparsity by going beyond the listing representation (i.e., they only store tuples with non-zero probability). In particular, the goal in these aforementioned works is to represent the factors compactly via \emph{algebraic} structures. 
%One of the earlier such representations was the Algebraic Decision Diagrams (ADDs) from~\cite{add}. 
Among related representations, \emph{arithmetic circuits} or ACs~\cite{ijcai2005} %{\asa{TODO describe ACs, SDD and sum product networks}}%. The main idea here is to exploit factor sparsity by going beyond the listing representation (i.e. only store the tuples with non-zero probability). In particular, the goal in these aforementioned works is to represent the factors with a more compact \emph{algebraic} structures. One of the earlier such representation was the Algebraic Decision Diagrams (ADDs) from~\cite{add}. Among related representations, \emph{arithmetic circuits} or ACs~\cite{ijcai2005} are one of the most actively researched variants. %{\asa{TODO describe ACs, SDD and sum product networks}}
%The idea in this line of work is it first represent the function as a CNF formula and then to create arithmetic circuits (i.e. circuits that can use operations over a field instead of Boolean operations) equivalent to this CNF formula. The inference then happens directly in the AC representations. 
 are still a very active research area~\cite{darwiche_nips16}. At a very high level, these circuits work best when the PGM variables themselves are Boolean. However, the `compilation' process (to AC representation) becomes more expensive for larger domain sizes. \begin{newEdits}Although our algorithm also seeks to exploit factor sparsity, it works directly with the much simpler listing representation, making it much more efficient.\end{newEdits}

%The latter is the focus of our work and for PGMs with large domain sizes (and sparse factors), our algorithm that works directly with the much simpler listing representation\footnote{Arithmetic decision diagrams (ADDs) and sparse vectors can in theory be converted to a listing representation~\cite{faq}}  is much more efficient.
 
An emerging area \emph{lifted probabilistic inference}~\cite{braz05,milch08,kersting12}, exploits symmetric structures within graphs to speed up inference. Thus far, this has been undertaken primarily in the relational learning paradigm, whereas our current work is propositional.
  
Yet another stream runs along the lines of variable elimination~\cite{dechter1996,ZhangPoole1996}, which undertakes a sequential marginalization of variables to compute posterior probabilities. Another stream involves tree decomposition-based routines~\cite{jensen1990,kask2005,IJGP} where the original network is decomposed into a hypertree and inference is performed using a two-phase message passing routine on this decomposed tree. The runtime complexity of %a large number 
most of these algorithms is dictated by the \emph{treewidth} of the underlying graph. {\ourModel} improves on this class of algorithms, with its complexity bounded by a finer notion of \emph{fractional hypertree width}.

Finally, past work in PGMs has also focused on approximate inference~\cite{koller2009}; we believe that the advancements introduced in {\ourModel} could enhance existing approximate engines by, for instance, applying it to models in the emerging tractable learning paradigms~\cite{bekker2015} such as thin junction trees~\cite{thinjt}.
%for large graphs due to scalability issues in exact methods. We believe that the advancements introduced in this work can potentially enhance existing approximate inference engines, for instance, by applying it to models in the emerging tractable learning paradigm~\cite{bekker2015} such as thin junction trees~\cite{thinjt}.

\section{\ourModel: An Overview}
\label{sec:overview}

Below we (a) give a brief overview of the background concepts, (b) outline the theoretical algorithm and (c) identify the implementation challenges and present our solutions.
%The formal details can be found in the supplementary material.

\subsection{Background}

\begin{definition} \label{Definition:pgm}
A (discrete) \emph{\textbf{probabilistic graphical model}} can be defined by the triplet $\langle \mathcal{H}, \boldsymbol{D},\mathcal{K} \rangle$ where hypergraph $\mathcal{H} = (\mathcal{V},\mathcal{E})$ represents the underlying graphical structure (note $\mathcal{E}\subseteq 2^{\mathcal{V}}$). There are $n=|\mathcal{V}|$ discrete random variables on finite domains $\boldsymbol{D} = \{D(U): U \in \mathcal{V}\}$ and $m = |\mathcal{E}|$ factors $\mathcal{K} = \{\phi_e |e \in \mathcal{E}\}$, where each factor $\phi_e$ is a mapping: $\phi_e:\prod_{U\in e}D(U)\rightarrow \mathbb{R}_+$.
%\begin{equation*}
%\phi_e:\prod_{U\in e}D(U)\rightarrow \mathbb{R}_+.
%\end{equation*}
\end{definition}

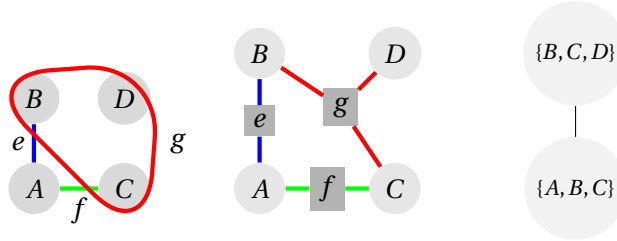
\begin{figure}[H]
%\begin{wrapfigure}{L}{0.5\textwidth}
\begin{center}
\hspace*{-1cm}
\begin{tikzpicture}[scale=0.6]

%%% The pGM query hypergraph
\node[circle, fill=gray!30] (A) at (0,0) {$A$};
\node[circle,  fill=gray!30] (B) at (0,2) {$B$};
\node[circle,  fill=gray!30] (D) at (2,2) {$D$};
\node[circle,  fill=gray!30] (C) at (2,0) {$C$};

\draw[ultra thick, blue] (A) -- (B); \node[left] at (0,1) {$e$};
\draw[ultra thick, green] (A) -- (C); \node[below] at (1,0) {$f$};
\draw[rounded corners=10mm, ultra thick, red] (-1.3,2.5) -- (2.8, 2.8) -- (2.5, -1.3) --cycle; \node[right] at (2.8,1) {$g$};

%%% The factor graph
\node[circle, fill=gray!20] (AFac) at (5,0) {$A$};
\node[circle,  fill=gray!20] (BFac) at (5,3) {$B$};
\node[circle,  fill=gray!20] (DFac) at (8,3) {$D$};
\node[circle,  fill=gray!20] (CFac) at (8,0) {$C$};

\node[rectangle, fill=gray!60] (e) at (5,1.5) {$e$};
\node[rectangle, fill=gray!60] (f) at (6.5,0) {$f$};
\node[rectangle, fill=gray!60] (g) at (6.8,1.8) {$g$};

\draw[ultra thick, blue] (AFac) -- (e) -- (BFac);
\draw[ultra thick, green] (AFac) -- (f) -- (CFac);
\draw[ultra thick, red] (BFac) -- (g) -- (CFac);
\draw[ultra thick, red] (g) -- (DFac);

%% The GHD

\node[circle, fill=gray!10] (Bone) at (12,0) {{\footnotesize $\{A,B,C\}$}};
\node[circle, fill=gray!10] (Btwo) at (12,3) {{\footnotesize $\{B,C,D\}$}};
\draw (Bone) -- (Btwo);
\end{tikzpicture}
\end{center}
\caption{The PGM query hypergraph (left), factor graph (middle) and a GHD for it (right). We use hypergraphs instead of factor graphs for notational ease.}
\label{fig:pgm}
%\end{wrapfigure}
\end{figure}

For instance Figure~\ref{fig:pgm} is a hypergraph representing a PGM with variables $\mathcal{V} = \{A, B, C, D\}$, edges $\mathcal{E} = \{e {=} (A, B)$, $f {=} (A, C)$, $g {=} (B, C, D)\}$ and factors $\mathcal{K}=\{\phi_{e}(A,B)$, $\phi_{f}(A,C)$, $\phi_{g}(B, C, D)\}$.

\begin{definition}
\label{defn:sparsity}
For any factor $\phi$, %(joint, input-factors or messages), 
the size of $\phi$ is its support size, i.e., the number of entries with non-zero probabilities. 
%denoted by $|\phi|$ is the number of entries with non-zero probabilities 
Storing only the non-zero entries (as well as their $\phi$ values) is called the \emph{listing representation} of $\phi$. \emph{Factor sparsity} is defined as $N/({\prod\limits_{U \in \phi} |D(U)|})$, where $N$ is the size of factor $\phi$. %When $\phi$ represents a \lq{}factor product\rq{} %joint probability distribution of a bag, 
%we call its factor sparsity as \emph{join table sparsity}.
\end{definition}

A typical inference task in PGMs is to compute the marginal estimates given by: $\forall F \subseteq \mathcal{V}$, $\boldsymbol{y}\in \prod\limits_{U\in F} D(U)$,
\begin{equation}\label{eq:marginal}
\phi_F(\boldsymbol{y}) = \frac{1}{Z} \sum\limits_{\boldsymbol{z} {\in} {\prod\limits_{U \in \mathcal{V}\backslash F}}D(U)} \prod\limits_{S \in \mathcal{E}} \phi_S(\boldsymbol{\mathrm{x}}_S), 
\end{equation}

$\text{where } \boldsymbol{\mathrm{x}} = (\boldsymbol{y}, \boldsymbol{z})$, $\boldsymbol{\mathrm{x}}_S$ denotes the projection of \boldsymbol{\mathrm{x}} onto the variables in $S$ and $Z$ is a normalization constant. Variable/factor marginals are a special case of (\ref{eq:marginal}); $F = \{U\}$ for $U \in \mathcal{V}$ for variable marginals and $F \in \mathcal{E}$ for factor marginals. 

Exact inference in PGMs is usually performed by propagating on a \emph{\textbf{generalized hypertree decomposition (GHD)}} of the underlying hypergraph $\mathcal{H}$. 

\begin{definition} \label{Definition:ghd}
A GHD of $\mathcal{H}=(\mathcal{V},\mathcal{E})$ is defined by a triple $ \langle T,\chi,\lambda \rangle$, where $T=(V(T), E(T))$ is a tree, $\chi: V(T) \rightarrow 2^{\mathcal{V}}$ is a function associating a set of vertices $\chi(v) \subseteq \mathcal{V}$ to each node $v$ of $T$, and $\lambda: V(T) \rightarrow 2^{\mathcal{E}}$ is a function associating a set of hyperedges to each node $v$ of $T$ such that the following properties hold (i) for each $e \in \mathcal{E}$, there is a node $v \in V(T)$ such that $e \subseteq \chi(v)$ and $e \in \lambda(v)$; and (ii) for each $t \in \mathcal{V}$, the set $\{v \in V(T)|t \in \chi(v)\}$ is connected in $T$.
\end{definition}

\subsubsection{ Existing GHD-based message passing algorithms}
As illustrated in Figure~\ref{fig:pgm}, a GHD can be thought of as a labeled (hyper)tree $T$, where sets assigned to each node in $T$ are called \emph{bags} of the hypertree.  %Each node of the tree $v$ is labeled; $\chi(v)$ describes which \emph{variables} are \lq\lq{}returned\rq\rq{} by the node $v$ %- this mapping gives us the variable marginals for all $U \in \chi(v)$. 
%The label $\lambda(v)$ captures the set of \emph{factors} that are combined into a joint probability estimate at this particular node.
% - this mapping gives us the factor marginals for all $e \in \lambda(v)$. The sets $\chi(v), v \in V(T)$ are called \emph{bags} of the tree-decomposition. 
%
Inference propagation on $T$ involves a two-pass \lq{}message-passing\rq{} algorithm~\cite{jensen1990}. %~\cite{jensen1990,shafer1990}.
%\iffalse %--NIPS 2017
In the first pass (message-up), \emph{messages} are propagated \lq{}up\rq{} from leaf (child) to root (parent). Subsequently in the second pass (message-down), they are propagated \lq{}down\rq{} from root (parent) to leaf (child).

\paragraph{} A \emph{message} $\phi\rq{}_{m_{v,u}}$ from node $v$ to node $u$ can be viewed as a marginal estimate where variables $U \not\in \chi(v) \cap \chi(u)$ are summed out of the factor product of node $v$, which is given by:
\begin{equation}\label{eq:joint}
\phi\rq{}_v = \prod_{e \in \lambda(v)} \phi_e \cdot \prod_{w \in \Gamma(v)} \phi\rq{}_{m_{w,v}},
\end{equation}
where $\Gamma(v)$ represents $\children(v)$ or $\parent(v)$ depending on the propagation direction. Upon completion of both propagation passes variable marginals for all $U \in \chi(v)$ can be retrieved using label $\chi(v)$, and factor marginals for all $e \in \lambda(v)$ can be retrieved from each node $v$ using the label $\lambda(v)$.

\subsection{Joins based theoretical inference algorithm}

The above GHD based messaging passing framework (also known as junction tree algorithms) forms the structure of \ourModel\ as outlined in Algorithm~\ref{jta} (in standard algorithms $R$ is always $0$, and in FAQ/AJAR %the theoretical algorithms in~\cite{faq,ajar} 
 $R$ is always $1$). % i.e. no 01-projections are selected for any of the bags).
To compute the GHD in Line~\ref{line:JI-ghd} we build a junction tree using the MinFill heuristic, rooting it arbitrarily and determine the parent-child relationship for every node. We assume that each input factor in $\mathcal{K}$ is assigned to a unique bag in the GHD. In particular, for every node $v$, $\alpha{(v)}$ denotes the factors in $\mathcal{K}$ assigned to it. 
\begin{algorithm} \label{junction_tree_algorithm}
\caption{\ourModel} \label{jta}
\begin{algorithmic}[1]
\small
\State{\textbf{Input:} A PGM $\mathcal{P} = (\mathcal{H}, D, \mathcal{K})$.}
\State{\textbf{Output:} Variable and Factor Marginals.}
\State{\text{Create a GHD $\mathcal{G} = ((V,E), \chi, \lambda)$ for $\mathcal{P}$.}} \Comment{Using MinFill variable ordering.}
\label{line:JI-ghd}
\State $R\gets$ \ourModelSampling$(\mathcal{G})$ \Comment{$R : V \to \{0, 1, 2\}$.}
%\State{$(\{\phi'_{v}, \phi'_{m_{v, \parent(v)}}\}_{v \in V}) \gets$ \text{\ourModelUp}$(\mathcal{G} , R)$.}
\State{$(\{\phi'_{v}\}_{v \in V}, \{\phi'_{m_{v, \parent(v)}}\}_{v \in V})\gets$ \text{\ourModelUp}$(\mathcal{G} , R)$}
\State{$\{\phi'_{v}\}_{v \in V}\gets \ourModelDown(\mathcal{G},  \{\phi'_{v}\}_{v \in V},\{\phi'_{m_{v, \parent(v)}}\}_{v \in V})$}
\State{\text{Compute variable and factor marginals from} $(\{\phi'_{v}\}_{v \in V})$}
\end{algorithmic}
\end{algorithm}

\begin{newEdits}
\subsubsection{Message-Up Phase}

\paragraph{Computation Within a Single Bag.} The upward pass propagates messages or marginalized factor products from leaf-nodes (child bags) to root-nodes (parent bags) along the hypertree. This involves computing factor products at every node (bag) of the hypertree. The runtime complexity of junction tree algorithms is dominated by these bag-wise factor product computations, and, it is here that \ourModel\ lends its core contribution: it uses a novel algorithm different from previously proposed exact inference algorithms to compute the factor products inside each bag. By exploiting the correspondence between computing database joins and computing factor marginals\footnote{If the probability values in the factor entries are Boolean, i.e., just 0 or 1, the factor product would reduce to a join.}, it uses worst-case optimal join algorithms (WCOJA) to compute the factor products \emph{within each bag}. The key idea behind these algorithms is that, unlike the traditional approach of computing factor products via a sequence of pairwise products, WCOJA undertake a multi-way product, i.e., the product of all relevant factors are computed simultaneously. Moreover, they work on a listing representation of the factors thereby exploiting the inherent factor sparsity of a PGM. The multi-way product algorithm used by {\ourModel} is outlined in Algorithm~\ref{nprr}, which is essentially from~\cite{ngo2}.

%\subsubsection{Description of \NPRR\ (Algorithm~\ref{nprr})}
\paragraph{} The inputs to Algorithm~\ref{nprr} are: a product flag that decides between the Multiway Product algorithm of~\cite{ngo2} and Algorithm~\ref{PP}, the set of variables $V$ whose product $\phi'$ needs to be computed, the corresponding factors $\mathcal{E}$, factor tables $\mathcal{K}$ and finally, the set of free variables $\mathcal{F}$. When the product flag is $2$, we run the traditional Pairwise Product algorithm. Otherwise, we run our algorithm whose description we provide here. We assume that $V$, $\mathcal{E}$, $\mathcal{K}$ and $\mathcal{F}$ are all sorted in input order. We now start with $V_{1}$ and find a value $y \in D(V_{1})$ that is present in all the factor tables of factors which contain $V_{1}$. We then fix this value as a potential candidate for $V_{1}$ in the factor product and continue the process for $V_{2}$ and so on. If we get a potential candidate from every variable in $V$, we have a vector that we add to the final factor product. On the other hand, if we don't get a potential candidate in a given variable say $V_{k}$, we backtrack and find a new potential candidate for $V_{k - 1}$ (i.e. a value that is present in all factor tables of factors containing $V_{k - 1}$) and so on (until $V_{1}$).  We obtain two outputs from this algorithm: the factor product $\phi'$ and its marginal on $\mathcal{F}$ denoted by $\phi'_{\mathcal{F}}$.

\begin{algorithm} \label{multi_fac_prod}
	\caption{\NPRR} \label{nprr}
	\begin{algorithmic}[1]
		\small
		\State{\textbf{Input:} Product Flag $P$, Variables $V$, Factors $\mathcal{E}$, Factor Tables $\mathcal{K}$ (as tries) and Free variables $\mathcal{F}$.} \Comment{All these are sorted in the input order.}
		\State{\textbf{Output:} Factor Product $\phi'$ and its marginal on $\mathcal{F}$, denoted by $\phi'_{\mathcal{F}}$.}
		\If{$P  = 2$}
			\State{$(\phi', \phi'_{\mathcal{F}}) \gets \PP(V, \mathcal{E}, \mathcal{K}, \mathcal{F})$} \Comment{Algorithm~\ref{PP}.} \\
			\qquad \Return{$(\phi', \phi'_{\mathcal{F}})$}
		\EndIf
		\State{Initialize a vector $\mathbf{x}$ of size $|V|$ to all $0$s.} \Comment{The entries of $\phi'$ will be added one-by-one as a vector ($\mathbf{x}$), probability pair.}
		\State{$x_1 \gets - \infty, k \gets 0$}
		\While{$x_1 < \infty$}
			\State{$y \gets x_{k + 1}$}
			\While{\text{(true)}}
				\ForAll{$S \in \mathcal{E}_{v} : S \cap V_{k + 1} \neq \emptyset$} \Comment{All factors that contain the variable $V_{k + 1}$.}
					\State{$y^{S}_{k + 1} \gets \min{\{x_{k + 1} | (x_{k + 1} > y) \wedge (x_{k + 1 } \in \mathcal{K}_{S_{k + 1}}) \}}$} \Comment{Here, $\mathcal{K}_{S}$ is the factor table corresponding to $S$. Among all $x_{k + 1} \in D(V_{k + 1}): x_{k + 1} > y$ having at least one entry in $\mathcal{K}_{S}$, we pick the smallest one (using \textit{galloping}).}
				\EndFor
				\State{$y \gets \max_{S}\{y^{S}_{k +1}\}$}
				\If{$(y = \min_{S}\{y^{S}_{k + 1}\})$} \Comment{Check if the value $y$ is present in all factor tables of factors that contain $V_{k + 1}$.}
						\State{$x_{k + 1} \gets y$}
						\State{\textbf{break}}
				\Else{}
					\State{$y = y - 1$}
				\EndIf
			\EndWhile
\If{$(x_{k + 1} \neq \infty)$}
	\If{$k + 1 < |V|$}
		\State{$k \gets k + 1$} \Comment{The value for variable $V_{k}$ is fixed as $x_{k}$ and we move on to the next variable.}
					\State{$x_{k + 1} \gets - \infty$}
				\Else
					\State{$\phi' \gets \phi' \cup \{(\mathbf{x}, \prod_{S \in \mathcal{E}} \mathcal{K}_{S}(x_{S}))\}$}
				\EndIf
			\Else{}
				\If{$k > 0$}
					\State{$k \gets k - 1$} \Comment{We don't have any potential candidates for $V_{k + 1}$ and we backtrack.}
				\EndIf
		\EndIf
		\EndWhile
\If{$\mathcal{F} \neq \emptyset$}
\State{$\phi'_{\mathcal{F}} \gets \sum_{q \in V \setminus \mathcal{F}} \phi'$} \Comment{We marginalize out $\phi'$ on all the variables in $\mathcal{F}$, resulting in a set of vector, probability pairs.}
\EndIf
\State{\Return{$(\phi', \phi'_{\mathcal{F}})$}}
\end{algorithmic}
\end{algorithm}

\begin{algorithm} \label{pairwise_prod}
	\caption{\PP} \label{PP}
	\begin{algorithmic}[1]
			\State{\textbf{Input:} Variables $V$, Factors $\mathcal{E}$, Factor Tables $\mathcal{K}$, Free variables $\mathcal{F}$.} \Comment{All these are sorted in the input order.}
			\State{\textbf{Output:} Factor Product $\phi'$ and its marginal on $\mathcal{F}$, denoted by $\phi'_{\mathcal{F}}$}
			\State{$\phi' \gets \emptyset$}
			\ForAll{$e \in \mathcal{E}$} 
				\If{$\phi' == \emptyset$} 
					\State{$\phi' \gets \mathcal{K}_{e}$}
				\Else
					\State{$\phi' \gets \phi' \cdot \mathcal{K}_{e}$} \Comment{This step is computed using LibDAI's API for pairwise product.}
				\EndIf
			\EndFor
			\State{$\phi'_{\mathcal{F}} = \sum_{q \in V \setminus \mathcal{F}} \phi'$} \Comment{This is step is computed using LibDAI's API for marginalizing out variables from a Factor/Cluster Product.}
			\State{\Return{$(\phi', \phi'_{\mathcal{F}})$}}
	\end{algorithmic}
\end{algorithm}

\begin{example}\label{eg:bag1}
Let us look at the \emph{triangle query} $\phi\rq{}(A, B, C) = \phi(A, B) \cdot \phi(B, C) \cdot \phi(C, A)$ for a bag. For simplicity, let  $|D(U)| = D$ for all $U \in \{A, B, C\}$ and $|\phi_e| = N \leq D^2$ for all $e \in \{(A,B), (B,C), (C,A)\}$.\begin{newEdits}Standard algorithms compute this query in time $O(D^3)$. %as they completely ignore factor-sparsity. 
On the other hand, we obtain a runtime bound of $O(N^{\frac{3}{2}})$, a bound that becomes especially useful in the presence of factor sparsity i.e., $N < D^2$ %We illustrate this point using one of the simulated datasets (Section~\ref{sec:exp_setup}) that we generate using SNAP~\cite{snapnets}: here, $D = 4039$ making $D^3 \approx 6\cdot (10^{10}) $, and $N = 88234$ making $N^{\frac{3}{2}} \approx 2\cdot (10^7)$; a 3 order magnitude gain. %However, 
(in the worst case $N = D^2$, we recover the bound $O((D^2)^{3/2}) = O(D^3)$).\end{newEdits}
\end{example}

\paragraph{Computation Across Multiple Bags.} Another significant difference stems from the \lq{}factors\rq{} that are included in the bag-wise product computation~\cite{dechter08}. Standard algorithms process (a) the input factors mapped to the bag and (b) the messages received by the bag from its children. In addition to the above two, \ourModel\ also includes factors called \lq{}01-projections\rq{} that are not originally mapped to the bag, but have non-trivial intersections with the variables in the bag. In other words, such factors are computed \emph{across multiple bags} in the hypertree. In the presence of sparsity, this \lq{}look-ahead\rq{} maneuver helps prune unproductive entries from factor products early on. Further, using 01-projections is central to realizing the asymptotically better bounds in FAQ/AJAR (see Section~\ref{sec:runtime}).

\begin{example}\label{eg:bag2}
Consider the bag $\{A, B, C\}$ of the GHD in Figure~\ref{fig:pgm} (where assume $F=\{B,C\}$) representing a product over two factors, $\phi\rq{}(a, b, c) = \phi_{e}(a, b) \cdot \phi_{f}(a, c)$. The \lq{}up\rq{} pass would propagate $\phi\rq{}({b, c)} = \sum_a \phi\rq{}(a, b, c)$, a marginal on $\{B,C\}$ to the root bag $\{B, C, D\}$. The worst-case output size of this factor product is given by $|\phi_{e}(a, b) \cdot \phi_{f}(a, c)| \leq |\phi_{e}(a, b)| \cdot |\phi_{f}(a, c)| = N^2$ and consequently the runtime bound is $O(N^2)$. However, we are able to obtain a much better bound by also utilizing factor $\phi_{g}(B, C, D)$ in the product computation (as both $B$ and $C$ participate in this factor), $\psi (A, B, C) = \phi_{e}(A, B) \cdot \phi_{f}(A, C) \cdot \phi_{01}(B, C)$, where $\phi_{01}(b, c) = 1$, $\exists$ d s.t. $\phi_{g}(b,c,d) \neq 0 \text{ and } 0 \text{ otherwise }$. By employing this 01-projection in our factor product computation, we can get a bound of $N^{3/2}$ on the output size (recall Example~\ref{eg:bag1}) and consequently a runtime bound of $O(N^{3/2})$ for the above query (detailed illustration and formal definition of 01-projections are provided later in Appendix~\ref{01projections}). 
\end{example}

\paragraph{} The upward message propagation in \ourModel\ is outlined in Algorithm~\ref{up}. In FAQ/AJAR, the condition in Line~\ref{line:JI-up-chk-R} is always satisfied while in traditional algorithms, the condition is never satisfied. Further, traditional algorithms in lines~\ref{line:call-nprr-1} and~\ref{line:call-nprr-2} use a pairwise product algorithm, which is asymptotically slower than WCOJA.

\begin{algorithm} \label{message_up}
	\caption{\ourModelUp} \label{up}
	\begin{algorithmic}[1]
		\small
		\State{\textbf{Input:} GHD $\mathcal{G} = ((V,E), \chi, \lambda)$ and the map $R$.}%i.e. the output of Algorithm~\ref{sampAlgo}.}
		\State{\textbf{Output:} Factor Products $\{\phi'_{v}\}_{v \in V}$ and Up Messages $\{\phi'_{m_{v, \parent{(v)}}}\}_{v \in V}$ both as tries.}
		\ForAll{\text{nodes} $v \in V$}: \Comment{This is done in a level-order traversal from leaves-to-root.}
		\State{$\mathcal{E}_{v}\gets \lambda(v), \mathcal{K}_{v} \gets \{\phi_{e}: e \in \lambda(v)\}$}  \Comment{Initialize the PGM Query corresponding to $v$'s Factor Product.}
		%\ForAll{$e \in \lambda{(v)}$} \Comment{We add all the factors in $v$ to $\mathcal{K}_{v}$.}
		%\State{$\mathcal{E}_{v} \gets \mathcal{E}_{v} \cup \{\lambda{(v)}\}, \mathcal{K}_{v} \gets \mathcal{K}_{v} \cup \{\phi_{e}\}_{e \in \lambda{(v)}}$} \Comment{We add all the factors in $v$ to $\mathcal{K}_{v}$.}
		%\EndFor
		\ForAll{$w \in \children{(v)}$} \Comment{We add all the messages sent to $v$ from its children.}
		\State{$\mathcal{E}_{v} \gets \mathcal{E}_{v} \cup \{\chi{(v)} \cap \chi{(w)}\}, \mathcal{K}_{v} \gets \mathcal{K}_{v} \cup \{\phi'_{m_{w, v}}\}$}
		\EndFor
		\If{$R(v) = 1$} \Comment{We include the $0/1$ projections while computing the Factor Product for $v$. }
		\label{line:JI-up-chk-R}
		\ForAll{$e \in \mathcal{E} \setminus \lambda{(v)}$} 
		\If{$e \cap \chi{(v)} \neq \emptyset$} 
		\State{$\mathcal{E}_{v} \gets \mathcal{E}_{v} \cup \{e \cap \chi{(v)}\}, \mathcal{K}_{v} \gets \mathcal{K}_{v} \cup \{\phi_{e / \chi{(v)}}\}$}
		\EndIf
		\EndFor
		%\State{$\phi_{v} \gets \prod_{e \in \lambda{(v)}} \phi_{e} \cdot \prod_{c \in children(v)} \phi'_{c, v} \cdot \prod_{e \in \mathcal{E} \setminus \lambda{(v)}, e \cap \chi{(v)} \neq \emptyset} \phi_{e \cap \chi{(v)}}$} \Comment{We include the $0/1$ projections for the bag along with the input factors and the received messages from its children to compute the joint distribution.}
		%\Else 
		%\State{$\phi_{v} \gets \prod_{e \in \lambda{(v)}} \phi_{e} \cdot \prod_{c \in children(v)} \phi'_{c, v}$}
		%\Comment{We include only the input factors and the received me	ssages from its children to compute the joint distribution.}
		\EndIf
		\If{$v$\text{ is not a root}}
		%\State{Let $\phi'_{m_{u, v}}$ be $\sum_{\mathbf{z} \in \prod_{U\in \chi{(v)} \setminus \chi{(u)}} D(U)} \phi'_{v}, v \in \Gamma{(u)}$.}
		\State{Let $u = \parent(v)$, $\phi'_{m_{v, u}}$ be $\sum_{q \in \chi{(v)} \setminus \chi{(u)}} \phi'_{v}$ and $\mathcal{F}_{v} = \chi{(v)} \cap \chi{(u)}$.}
		%\State{$\phi'_{m_{u, v}} = \sum_{q \in \chi{(v)} \setminus \chi{(u)}} \phi'_{v}$ \text{where }$v \in \Gamma(u)$.}
		\State{$(\phi'_{v}, \phi'_{m_{v, u}}) \gets \NPRR(R_{v}, \chi{(v)}, \mathcal{E}_{v}, \mathcal{K}_{v}, \mathcal{F}_{v})$} 
		%\State{\text{Compute (as tries) }$\phi'_{v}$\text{ and }$\phi'_{m_{u, v}}$\text{ with }$\NPRR(\mathcal{E}_{v}, \mathcal{K}_{v}, 1)$.}
		\label{line:call-nprr-1}
		\Else{ $\phi'_{v} \gets \NPRR(R_{v}, \chi{(v)}, \mathcal{E}_{v}, \mathcal{K}_{v}, \chi{(v)})$}
		%\State{$(\phi'_{v}) \gets \NPRR(\chi{(v)}, \mathcal{E}_{v}, \mathcal{K}_{v}, \emptyset)$}
		%\State{\text{Compute (as tries) } $\phi'_{v}$\text{ with }$\NPRR(\mathcal{E}_{v}, \mathcal{K}_{v}, 0)$.}
		\label{line:call-nprr-2}
		\EndIf
		\EndFor
		\State{\Return{$\left(\{\phi'_{v}\}_{v \in V}, \{\phi'_{m_{v, \parent{(v)}}}\}_{v \in V}\right)$}}
	\end{algorithmic}
\end{algorithm}
%\subsubsection{01 projections}\label{01project2}
%Using $0/1$-projections is central to realizing asymptotically better bounds in FAQ/AJAR (see Section~\ref{sec:runtime}). Moreover, this gain is accentuated when we deal with sparsity: the key idea being to exploit the \emph{sparsity of all the input factors} and not just those encompassed by the query corresponding to the current bag. See the supplementary material for a more detailed explanation. 
\subsubsection{Message-Down phase}
\begin{newEdits}
Since the Message-Down phase (Algorithm~\ref{down}) involves updating factor products for each bag (except the root) using down messages, we perform two in-place $\hash$s for each such bag: first, between the up and down messages sent/received by the bag, and then, between the result of the previous step with the bag's factor product. A detailed description on how to compute an in-place $\hash$ follows.

\paragraph{} Consider any two factors $e_{1}$ and $e_{2}$, whose corresponding factor tables are denoted by $\phi_{e_1}$ and $\phi_{e_2}$ respectively. (Note that cluster products/messages can be treated as factor tables as well.) Recall that the factor tables are stored as a list of $\langle$ index, probability $\rangle$ pairs. We start by assuming that one of $\phi_{e_1}$ or $\phi_{e_2}$ is stored as a hash-table with index as key and probability as value respectively. \footnote{The up/down messages are stored as hash tables.} Without loss of generality, let's assume that $\phi_{e_1}$ is stored as a hash table. Our goal is to compute $\phi_{e_1} \phi_{e_2}$ in-place. To this end, we iterate through each entry in $\phi_{e_2}$ and probe the corresponding index in $\phi_{e_1}$'s hash table. If it is present, then we multiply the probabilities and store the result in the corresponding entry in $\phi_{e_2}$. Otherwise, we discard the entry from $\phi_{e_2}$. It follows that by the end of this procedure, we would have computed $\phi_{e_1} \cdot \phi_{e_2}$ and the result is stored in $\phi_{e_2}$. Note that probing in the Hash Table is an amortized constant time operation and entry removal in $\phi_{e_2}$ can be done in constant time as well, concluding that the time complexity of our algorithm is $O(|\phi_{e_2}|)$. 

%Since the Message-Down phase (Algorithm~\ref{down}) involves updating factor products for each bag (except the root) using down messages, we perform two in-place \emph{Hash Products} (see Section~\ref{inplacehash} in the Supplementary Material) for each such bag-- first, between the up and down message sent/received by the bag. Then, we multiply the result of the previous step with the bag's factor product.
\end{newEdits}
\begin{algorithm} \label{message_down}
\caption{\ourModelDown} \label{down}
\begin{algorithmic}[1]
\small
\State{\textbf{Input:} GHD $\mathcal{G} = ((V,E), \chi, \lambda)$. Factor Products $\{\phi'_{v}\}_{v \in V}$ and Up Messages $\{\phi'_{m_{v, \parent(v)}}\}_{v \in V}$ converted to listing representation from tries.}
\State{\textbf{Output:} Final Factor Products $\{\phi'_{v}\}_{v \in V}$.}
\State{Set $f[v] \gets 0$ for every $v \in V$.}
\ForAll{\text{nodes} $v \in V$}: \Comment{This is done as a level-order traversal from root-to-leaf.}
\ForAll{$w \in \children{(v)} :  f[w] = 0$}
\State{$\phi''_{m_{v, w}} \gets \sum_{q \in \chi{(v)} \setminus \chi{(w)}} \phi'_{v}$} \Comment{Compute the Down Message from $v$ to $w$ by summing out variables not in $\chi(w)$.}
\State{$ \phi'_{m_{v, w}} \gets \hash(\phi'_{m_{w, v}}, \phi''_{m_{v, w}})$} \Comment{We divide the Down Message by its corresponding Up Message.}
\EndFor
\ForAll{$w \in \children{(v)} :  f[w] = 0$}
\State{$ \phi'_{w} \gets \hash(\phi'_{w}, \phi^{'}_{m_{v, w}})$} \Comment{We multiply $w$'s cluster product with $\phi'_{m_{v, w}}$.}
\State{$f[w] \gets 1$.}
\EndFor
\EndFor
\State{\Return{$\{\phi'_{v}\}_{v \in V}$}}
\end{algorithmic}
\end{algorithm}%\label{down}

\paragraph{} Summing up, our inference algorithm is a standard tree propagation algorithm with two modifications: (1) We adapt WCOJA to compute factor-products in the bags, and
(2) We modify the factor products in bags using 01-projections.

\subsubsection{Runtime complexity}
\label{sec:runtime}
With the above steps we obtain a two-pass inference routine whose runtime bound  follows from a recent improved bound (AGM) on size of factor product~\cite{ngo2}. For a hypergraph $\mathcal{H} = (\mathcal{V},\mathcal{E})$, let $B \subseteq \mathcal{V}$ be any subset of vertices and let $\boldsymbol{x} \in \mathbb{R}^{|\mathcal{E}|}$ be a vector indexed by edges, such that $\boldsymbol{x}^*(B) = {(x^*_S(B))}_{S \in \mathcal{E}} $ be the optimal solution to the linear program
\begin{eqnarray}\label{eq:lpAGM}
\text{min}  && \sum_{S \in \mathcal{E}} x_S \text{log}_2 |\phi_S|\\ 
\text{s.t.} && \sum_{S:v \in S} x_S \geq 1, \forall v \in B; x_S \geq 0, \forall S \in \mathcal{E}.% \\
%\text{} && x_S \geq 0, \forall S \in \mathcal{E}.
\end{eqnarray}
%\begin{equation*}
%\text{min}  \sum_{S \in \mathcal{E}} x_S \text{log}_2 |\phi_s|, \text{s.t.} \sum_{S:v \in S} x_S \geq 1, \forall v \in B 
%\end{equation*}
%\begin{equation*}
%
%\end{equation*}

Then, the quantity %$ \text {AGM}_{\mathcal{H}} (B):= \prod_{S \in \mathcal{E}} |\phi_S|^{x^*_S}$
\begin{eqnarray*}
\text {AGM}_{\mathcal{H}} (B) &&  := \prod_{S \in \mathcal{E}} |\phi_S|^{x^*_S(B)}
\end{eqnarray*}
is called the AGM-{\em bound} of B using edges in $\mathcal{H}$. The WCOJA used in {\ourModel} meets the AGM-{\em bound} thus giving our GHD  (Definition~\ref{Definition:ghd}) based algorithm a runtime complexity of
\begin{equation}\label{agm_sum}
\sum_{v' \in V(T)}  \text {AGM}_{\mathcal{H}} (\chi(v')). %= \sum_{v' \in V(T)}  \prod_{S' \in \mathcal{E}'} |\phi_{S'}|^{x^*_{S'}},
\end{equation}
%where $\mathcal{E}'$ denotes the edges induced on $\mathcal{H}$ by variables in $\chi(v')$.

\paragraph{} Upper-bounding each $|\phi_S|$  by $N = \max_{S \in \mathcal{E}} |\phi_{S}|$ in the above equation and maximizing (instead of summing) over all bags in the GHD gives us an asymptotic bound of $N^{\fhtw(T)}$. $\fhtw(T)$ is guaranteed to be smaller than $\htw(T)$ or $\tw(T)$ (hypertree width or tree width) for the same GHD, giving us the best known theoretical bounds for exact inference in PGMs. (Formal definitions for $\fhtw(T), \htw(T), \tw(T)$ can be found in Appendix~\ref{ghdnotionsSupp}.)
%and derivation of the runtime complexity can be found in the supplementary material).

\begin{example}
In Example~\ref{eg:bag2}, the original hypergraph structure, $\phi\rq{}(A, B, C)$,  consisted of only two edges $(A, B)$ and $(A, C)$. The AGM-{\em bound} from the optimal solution $\boldsymbol{x}^* = \{1, 1\}$ translates to a bound of $N^2$. However, by employing 01-projections the %modified query presents a hypergraph with 
induced hypergraph, $\psi (A, B, C)$, has three edges $(A, B)$, $(A, C)$ and $(B, C)$ and the optimal solution $\boldsymbol{x}^* = \{1/2, 1/2, 1/2\}$ gives us an asymptotically better bound of $N^{3/2}$. For the GHD T, $\fhtw(T) = 3/2$,  $\htw(T) = \tw(T) = 2$.
\end{example}

\paragraph{} Note that (\ref{agm_sum}) gives a finer grained measure for our runtime: $\sum_{v' \in V(T)} \prod_{S \in \mathcal{E}} |\phi_S|^{x^*_S(\chi(v'))}$. Recall that the asymptotic bounds for $\tw$ based algorithms is given by $D^{\tw}$, where $D = \max_{U \in \mathcal{V}} |D(U)|$. However, a more realistic measure here would be $\rho = \sum_{v' \in V(T)} \prod_{U \in \chi(v')}| D(U)|$. This gives us a finer grained ratio (as compared to~\cite{dechter08}) to evaluate {\ourModel} against classical engines:

\begin{equation}\label{eqRJ}
R_J = \frac{\sum_{v' \in V(T)} \prod_{S \in \mathcal{E}} |\phi_S|^{x^*_S(\chi(v'))}}{\rho}.
\end{equation}

%Upper bounding $|\phi_S|$ with $N$ and $|D(U)|$ with $D$ and maximising (instead of summing) the quantities in the numerator and denominator of (\ref{eqRJ}) will give us:
Replacing the numerator with the upper bound  $N^{fhtw}$ and the denominator with the upper bound $D^{\tw}$ in (\ref{eqRJ}) gives us
\begin{equation}\label{eqRD}
R_D = \frac{N^{\fhtw}}{D^{\tw}}.
\end{equation}
This ratio is analogous to the one in~\cite{dechter08}, which was based on hypertree width:
\[ R = \log_{10}\left(\frac{N^{\htw^*}}{D^{tw}}\right).\]
Since computing $\htw$ is NP-Hard,~\cite{dechter08} used an approximation for it (denoted by $\htw^*$). In our measure $R_{D}$ , we overcome this issue by using $\fhtw$ over $\htw$. Using $\fhtw$ offers two significant advantages -- one, it is a more fine-grained measure (see~\eqref{eq:fhtwinequality}) and two, $\fhtw$ is polynomially computable (basically, solve the LP from~\eqref{eq:lpAGM}). We compute the numerator in $R_{D}$ as described in Appendix~\ref{ghdnotionsSupp}. We show in the Experiments Section that $R_{J}$ is a better predictor than $R_{D}$ on most bands since it exploits the fact that factors/factor tables and variables in a bag can have different sizes and domain sizes respectively.
%$R_D$ is an analogous measure for our algorithm (the ratio measure in~\cite{dechter08} is based on hyper-tree width: $\log_{10}(N^{htw}/D^{tw})$).

%$R_D$ is an analogous measure for our algorithm, however strictly speaking the ratio measure in~\cite{dechter08} is based on hyper-tree width: $\log_{10}(N^{htw}/D^{tw})$.

\subsection{Challenges and Our Solutions}

\subsubsection{Factor Representations}\label{Sec:factorRep}
Consistent with WCOJA, we use a \emph{listing representation}~\cite{Darwiche96,LarkinDechter03} to store data, i.e., we only store factor entries with non-zero probabilities. It has been shown that tries are sufficient for theoretical bounds\footnote{A trie is a multi-level data structure where each factor tuple corresponds to a unique path from root to leaf and the probability value associated with each tuple are stored in the leaf.}. While computing the factor product of a bag, Algorithm~\ref{up} adopts a back-tracking based search routine over multiple tries (\NPRR). However, \lq{}multi-level\rq{} tries impose considerable random access costs during the back-tracking search; they are also unsuitable for message propagation. We resolve these two problems via two novel factor-representations.

\paragraph{Within Bag Computation.} First, we introduce \lq{}level-order\rq{} tries for the back-tracking search in Algorithm~\ref{up}. Essentially, we \emph{flatten} the trie into a single-level contiguous block and redesign the search to mimic the original multi-level traversal. This re-design enabled us to simultaneously (i) exploit the compact storage of tries for sparse factors and (ii) the caching advantages of contiguous memory blocks. Initial experimental runs showed a gain of $\sim$10x in runtime using level-order tries. 

%\begin{newEdits}
\paragraph{Message Propagation Between Bags.} Second, in addition to storing each factor as a trie, we also store it as a list of $\langle \text{index value, probability} \rangle$ pairs where each factor tuple is converted into a single number: their index value. We use two variants (i) which stores only `reverse' indices, i.e., indices computed in reverse variable order and (ii) which stores forward and reverse indexes (for representing intermediary messages). These multiple representations enable us to optimize Algorithms~\ref{up} and ~\ref{down}: the reverse index enables efficient construction of tries in the up-phase and in decoding message entries over all children in a single pass in the down-phase. Moreover, the reverse indices of the up-messages act as placeholders for down-messages, enabling the re-use of data-structures. %(i.e., we don't have to create an explicit structure for the downward messages).
  Finally, the forward indexes are used while merging down-messages with cluster products, thus reducing redundant decoding/encoding steps.

\paragraph{01-projections.} To obtain the asymptotically better theoretical bounds, FAQ/AJAR uses 01-projections for all bag-wise computations. However, in practice they impose significant computational costs since interconnected factors generate a large number of projections per bag; building and maintaining these are costly. Firstly, since the utility of a 01-projection lies in its sparseness we filter out dense projections at each bag-wise computation. Secondly, instead of pre-computing all projections, we compute projections on the fly and amortize it (cache-and-reuse) over subsequent computations. 
\end{newEdits}

\begin{algorithm} 
%Let \alpha be a parameter that might need to get tuned. Now for any dataset, sort the clusters by product of domain sizes and pick the top k, where k is picked so that the sum of their product of domain sizes is <= \alpha*(the total sum of product of domain sizes): i.e. your code will have to figure out k. Then run both JI and libDAI on all of these k clusters (while making sure to cutoff JI if it takes more than time libDAI: if need be you can put in a factor here too) and make the decision based on these k runs.
 %The idea is to keep \alpha small: say <= 10%. Theoretically, if on a dataset libDAI is faster (and assuming the above sampling does indeed pick libDAI) then overall slowdown wrt libDAI is (1+\alpha). However, in the case where JI is much faster your runtime will be \alpha*libDAI + JI. So the latter would start hurting if JI is more than (1/\alpha) times faster.
\caption{\ourModelSampling}
\label{hybridAlgo}
\begin{algorithmic}[1]
\small
\State{\textbf{Input:} GHD $\mathcal{G} = ((V,E), \chi, \lambda)$, Sum of product of domains $\rho$.}
\State{\textbf{Output:} A map $R$ that maps every node $v \in V$ to $0$ $\left(\text{\NPRR \space w/o }  0/1\right)$, $1$ $\left(\text{\NPRR}~ 0/1\right)$ or $2$ $\left(\text{\PP}\right)$.} 
\State{Initialize $R_{v}$ to $-1$ for every $v \in V$.}
\If{$\rho \le 10^{9}$}
\ForAll{\text{nodes} $v\in V$ with. $|\alpha{(v)}| \ge 1$}: \Comment{This traversal is done in the order of their product of domain sizes, highest to lowest.}
\State{$\mathcal{E}_{v} \gets \alpha{(v)}, \mathcal{K}_{v}\gets \{\phi_e : e \in \alpha{(v)}\}$}
\State{$T_{0} \gets \text{Time taken to run } \NPRR\left (0, \chi{(v)}, \mathcal{E}_{v}, \mathcal{K}_{v}, \chi(v) \right)$}
\State{$\mathcal{E'}_{v} \gets \{S\cap \chi(v)|S\in E\}, \mathcal{K'}_{v}\gets \{\phi_e : e \in \mathcal{E'}_{v}\}$}
\State{$T_{1} \gets\text{Time taken to run } \NPRR \left(0, \chi{(v)}, \mathcal{E'}_{v}, \mathcal{K'}_{v}, \chi(v)\right)$}
\State{$T_{2} \gets \text{Time taken to run }  \NPRR \left(2, \chi{(v)}, \mathcal{E}_{v}, \mathcal{K}_{v}, \chi(v)\right)$}
\State{$R_{v} \gets i$, where $\min{(T_{0}, T_{1}, T_{2})} = T_{i}$.}
\ForAll{$w \in \subtree{(v)} : R_{w} = -1$} \Comment{Propagate the decision obtained for $v$ to all unassigned nodes in its subtree.}
\State{$R_{w} \gets R_{v}$}
\EndFor
\EndFor
\Else 
\ForAll{\text{nodes} $v \in V$}
\State{$R_{v} \gets \random{(0, 1)}$}
\EndFor
\EndIf
\State{\Return{R}}
%} \footnote{Note that even after this procedure, there could still exist nodes $u$ in $V$ with $R_{u} = -1$. To address this, we repeat steps $10-14$ for all roots $v \in V$ with $R_{v} = -1$.}
\end{algorithmic}
\end{algorithm}

\begin{newEdits}
\subsubsection{Hybrid Architecture}
In Bands 5 and 6 (Figure~\ref{introTab}), libDAI's pairwise product implementation demonstrates distinct advantages over {\ourModel}'s multi-way product. We explore the feasibility of leveraging the respective advantages of both these strategies in a new \{HY\}brid \{J\}oin \{AR\}chitecture (HYJAR). To build such a system, we use the native structure of {\ourModel} and import only the pairwise-product functionality from libDAI (we do not integrate the entire engine). Given the high costs of switching between the data-structures required for {\ourModel} and libDAI, the main challenge here was to devise a system that not only optimally chooses between the strategies per bag, but at the same time minimizes the switches between bags.  We overcome this challenge by introducing a deterministic heuristic (Algorithm~\ref{hybridAlgo}) that decides the optimal strategy ({\ourModel} (with or without 01-projections) or {\PP}) for each bag $v$ in the GHD that has at least one input factor assigned to it (i.e. $\alpha(v)\ge 1$).\footnote{Recall our earlier assumption that each factor table is assigned to a unique bag. As a result, not many bags are chosen in this process. Further we ignore the incoming messages for a bag $v$ when deciding on $R_{v}$, making this decision faster.} We then propagate this decision along the subtree of $v$, until it reaches a bag that was already assigned a decision. To decide the order of preference, we consider bags $v$ in decreasing order of $\prod_{U\in \chi(v)}|D(U)|$, with the intuition being that the larger bags dominate the runtime of libDAI. A detailed description of Algorithm~\ref{hybridAlgo} follows.

\paragraph{} The inputs to our algorithm are a GHD $\mathcal{G} = ((V,E), \chi, \lambda)$ and the sum of product of domains $\rho$. Our goal is output a map $R$ that maps every node $v \in V$ to $0$, $1$ or $2$. In particular, $R_{v} = 0$ implies that we would be running {\ourModel} without any $0/1$ projections for that node in the message up phase. Similarly, $R_{v} = 1$ and $R_{v} = 2$ imply that we would be running {\ourModel} with $0/1$ projections and Pairwise Product (Algorithm~\ref{PP}) without any $0/1$ projections respectively. We first consider the case when $\rho \le 10^9$ -- we iterate through all nodes $v \in V$ with $|\alpha(v)| \ge 1$ (i.e. at least one input factor table assigned) in the decreasing order of their corresponding product of domain sizes. In particular, we run {\ourModel} and \textsf{Pairwise Product} without 0/1 projections (i.e. the first and third algorithms) with only pre-assigned factor tables determined by $\alpha{(v)}$. For the second algorithm, since we include 0/1 projections, we include all factor tables whose variables have a non-empty intersection with the union of variables in the pre-assigned factors in the current bag. Once these three runs are done, we assign the fastest strategy to node $v$. We then propagate this decision to all nodes in its subtree for which no decision has been made so far. However, even after this procedure, there could be some nodes, which don't have an $R$ value assigned. In order to address this issue, we repeat the same procedure for the root i.e., we make an arbitrary choice of strategy for it and then propagate this choice along the remaining GHD. Finally, for the case when $\rho > 10^9$, running {\ourModel} without 0/1 projections could turn out to be very expensive (since we don't consider the messages). Thus, for each cluster, we randomly choose between {\ourModel} without 0/1 projections and with 0/1 projections. Note that we don't run \textsf{Pairwise Product} for this case since LibDAI crashes when $\rho > 10^9$.

%unless a bag in the subtree was already assigned a decision. To decide the order of preference, we consider bags $v$ in decreasing order of $\prod_{U\in \chi(v)}|D(U)|$, with the intuition being that the larger bags dominate the runtime of libDAI.
%(See Subsection~\ref{hyjardescription} for a detailed description.)
\end{newEdits}

\section{Experimental Evaluation}
\begin{newEdits}
In this section, we empirically validate {\ourModel} and outline features that influence {\ourModel}\rq{}s performance. Specifically, we describe our empirical setup and on standard benchmarks, we (i) demonstrate the scope of {\ourModel} vis-a-vis state-of-the-art systems, (ii) document performance gains of the hybrid setting and (iii) evaluate our technical contributions.
\end{newEdits}

% evaluate Algorithm~\ref{sampAlgo} and, (iii) undertake finer-grained analysis to identify parametric settings that influence {\ourModel}\rq{}s performance.  %Finally, we document the performance gains one can achieve using an idealized \lq{}best-of-all-worlds\rq{} engine.

%present and validate our technical contributions, namely new factor-representations, 01-projections and our flexibility to undertake multiple inference tasks, and, finally (v) present a descriptive study demonstrating our theoretical bounds. %%(c) demonstrate the ability of 01-projections to leverage sparsity (c) present and validate our technical contributions, namely 01-projections and new factor-representations, and, finally (iv) demonstrate the efficacy of our model vis-a-vis state-of-the-art systems in real world settings and our flexibility to undertake multiple inference tasks. (ii) present a descriptive study demonstrating our theoretical bounds 

\subsection{Experimental Setup}\label{sec:exp_setup}
%We describe the datasets, comparison engines, metrics and experiment settings used to validate that our Inference Engine competes with the state-of-the-art in \yell{refer LibDAI and IJGP} sections.
 
%\subsubsection{Datasets}\label{sc:data}
\paragraph{Datasets.} To create a testbed that spans the full range of cases illustrated in Figure~\ref{introTab}, we sample from three publicly available benchmark datasets: UAI\rq{}06~\cite{uai2006}, PIC 2011 \cite{pascal11} and the BN Learns dataset~\cite{bnLearn} (which subsumes the IJCAI\rq{}05 networks~\cite{ijcai2005}). Our testbed contains 52 networks. For {\bandone}, we selected the CELAR subset from PIC 2011. From the UAI'06 benchmarks, we selected grids (BN\_30-41) for {\bandtwo}, iscas89 (BN\_47-68) for {\bandthree}, the Speech Recognition DBNs (BN\_20-25) for {\bandfour} and, iscas85 (BN\_42-46) for {\bandsix}. We left out networks CELAR-SUB4, BNs 65, 66 and 68 in Bands 1 to 3 because \ourModel~does not (yet) support the precision to compute very large indices ($> 2^{63} - 1$). Finally, we selected networks from BN Learns which had more than $30$ variables (i.e., we filtered out the smaller ones); these populated Bands 4, 5 and 6.

\begin{newEdits}

\paragraph{} In order to improve the tractability of some of the larger networks for exact inference (high $\rho$ cases), we randomly induce factor table sparsity. Further, to ensure that all final cluster tables are non-empty, for every factor $e \in \mathcal{E}$, we force the entry $(i)_{U \in e}$ in the corresponding factor table $\mathcal{K}_{e}$ for every $i \in [\min_{U \in e}$\\$|D(U)|]$. These sparsity levels are consistent with the ranges found in other networks in the benchmark. In particular, for the CELAR subsets, the induced median factor sparsity is at 20\% and 40\% (original median sparsity was 100\%). However, for the networks in Bands 2 and 3, we ensure that the induced median factor table sparsity remains close to the original value of 50\%. (Note that inducing sparsity to improve model tractability is a well-accepted procedure in many practical settings~\cite{songHanICLR18,LarkinDechter03}). 

\end{newEdits}
 %(details in supplementary material). 
%This produces significantly sparse factors that provides us a good testbed to present our inference framework.

%IJCAI\rq{}05~\cite{ijcai2005} and UAI\rq{}06~\cite{uai2006}. We chose these two datasets as they contain networks that lend themselves to exact-inference (graphs with bounded $tw$). We create our subset pool using the following filtering criteria: networks whose maximum domain size is greater than two $(D > 2)$ and whose median factor sparsity is less than $70\%$. Here factor sparsity is given by $N/({\prod\limits_{U \in \phi} D(U)})$, where $N$ is the size of factor $\phi$. 

%\subsubsection{Comparison Engines} 
\paragraph{Comparison Engines.} We compare {\ourModel} against three state-of-the-art systems: ACE~\cite{ijcai2005}, an engine that explicitly exploits determinism, and, libDAI~\cite{libDAI} and IJGP~\cite{IJGP}, two award winning systems in the UAI 2010 inference challenge. Since {\ourModel} is for exact inference, we compare against the exact inference \lq{}settings\rq{} in  these models: the \lq{}Junction Tree\rq{} algorithm in libDAI and the \lq{}Join-Tree\rq{} propagation in IJGP (since it computes exact marginals only when the join-graph is a tree). In particular, IJGP requires the \lq{}degree\rq{} parameter to be set high, which we set to the number of variables in the network (see Table~\ref{table_real_3}'s description for more details). For ACE, we follow the recommended settings outlined for the IJCAI networks and the standard settings for the others -- commands \lq{}compile\rq{} and \lq{}evaluate\rq{} to compile the AC and run inference respectively. The compile/run commands for these engines can be found in Appendix~\ref{compilerunc}.

%the \lq{}Junction Tree\rq{} algorithm in libDAI and the \lq{}Join-Tree\rq{} propagation in IJGP. IJGP computes exact marginals only when the join-graph is a tree. This requires the \lq{}degree\rq{} parameter to be set high, which we set to the number of variables in the network. For ACE, we follow the recommended settings outlined for the IJCAI networks (details in the supplementary material) and the standard settings for the others - commands \lq{}compile\rq{} and \lq{}evaluate\rq{} to compile the AC and run inference respectively.

%We use the recommended settings for ACE on IJCAI - commands \lq{}compile\rq{} and \lq{}evaluate\rq{} to compile the AC and run inference respectively.

%which requires the \lq{}degree\rq{} parameter to be set high. Hence we set the \lq{}degree\rq{} to the number of variables in the graph for all our experiments.

%\subsubsection{Inference queries}
\paragraph{Inference Queries.} The inference query we evaluate is the computation of (all) variable marginals.  We observed that while \ourModel, IJGP and ACE process evidence (requiring it to be input separately), libDAI does not. Further, \ourModel, IJGP and ACE perform SAT-based singleton consistency and treat the resulting variables as evidence, which again libDAI doesn't. Hence, in order to ensure a fair comparison, we incorporate the evidence and singleton-consistency directly into the input given to the engines. In particular, we remove both these types of variables from the input network (note that this could remove some factors from the query as well). We then provide the updated network resulting from this procedure to all four engines in their respective formats: libDAI (.fg), IJGP (.uai), ACE (.uai) and {\ourModel}. We compare our marginal outputs with these engines, with an error limit of $0.00001$.

\paragraph{Evaluation Metrics and Settings.} We evaluate the systems on the time taken to compute variable marg-\\inals. We repeat each experimental run $5$ times and report the average of the runs. Additionally, we set a timeout of $60$ minutes for our experimental runs. We would like to note here that ACE requires separate compilation of the arithmetic circuit representing the input network (a non-standard design). For a fair comparison with other engines with end-to-end computations, we report both (i) the total of compilation and inference times and (ii) only the inference time for ACE. We ran all our experiments on a Linux server (Ubuntu 14.04 LTS) with Intel Xeon E5-2640 v3 CPU @ 2.60GHz and 64 GB RAM. 

\paragraph{Error Codes in Table~\ref{table_real_3}.} \label{error-codes} We describe the error codes in Table~\ref{table_real_3} here. \lq{}T\rq{} denotes engine-time out (60 mins). LibDAI and IJGP crash on all benchmarks where $\rho > 10^{9}$, due to huge pre-memory allocation and this is denoted by \lq{}F\rq{}. For ACE, we observed that for benchmarks BN\_30-39 in {\bandtwo}, it compiles successfully but throws a runtime exception due to precision issues. We believe that this is due to large treewidth ($\ge 39$) on these datasets. We denote this by \lq{}E\rq{}. For IJGP, we observed that it does approximate inference in benchmarks Munin1 and BN\_43-46 respectively. In particular, we recorded its final treewidth using the MinFill ordering on all benchmarks and compared it with {\ourModel}'s and libDAI's treewidth (both using the MinFill ordering) respectively. We noticed that the final treewidth reported by IJGP was much smaller than the treewidth reported by {\ourModel} and libDAI. Recall that we preprocess evidence and SAT-based singleton consistency on these benchmarks and thus, we concluded that IJGP does Approximate Inference on these datasets (which we denote by \lq{}A\rq{}).
\subsection{Experimental Results}\label{sec:exp}

\begin{newEdits}
%TODO: Remove the c432
\begin{table*}
	\centering
	%\resizebox{0.7\textwidth}{!}{
	%\begin{minipage}{0.7\textwidth}
	\caption{{ \small Benchmark Comparisons: The first column denotes the range of $\rho$, followed the band of the datasets (see Figure~\ref{introTab}) and the dataset name. The fourth column denotes the number of variables/factors, followed by $R_{J}$ and $R_{D}$. We report three runtimes for {\ourModel}: without 0/1 projections, with all 0/1 projections and HYJAR, followed by our comparison engines -- LibDAI, IJGP and ACE (Total Time and Inference Time). (All runtimes are in seconds.) Further, we report the median and mean sparsity for every dataset, followed by fractional hypertree width (fhtw) and tree-width (tw) (computed for the same GHD). The fractional hypertreewidth ($\fhtw$) numbers were generated by solving the linear program~\eqref{eq:lpAGM} using Google OR-Tools~\cite{orsolver}. Finally, we report the maximum domain value (D) and maximum factor table (non-zero) entry size (N). \lq{}T\rq{} denotes engine-time out (60 mins). Engine crash due to huge pre-memory allocation is denoted by \lq{}F\rq{}.  Runtime exceptions due to precision issues is denoted by \lq{}E\rq{}. When an engine does Approximate Inference due to large treewidth, we denote it by \lq{}A\rq{}). See Section 4.1 for a detailed description of these error-codes.}}
	\label{table_real_3}
	%\resizebox{0.6\textwidth}{!}{
	\resizebox{1.0\textwidth}{!}{
	\begin{tabular}{|c|cccccccccccccccc|}

	$\rho$ & Band & Dataset & Var/Factors & $R_{J}$ & $R_{D}$ & \multicolumn{2}{c}{\ourModel}  & HYJAR & libDAI & IJGP & \multicolumn{2}{c}{ACE}  & Sparsity (in \%) & fhtw & tw & D/N \\
	\hline
	& & & & & & w/o 0/1 & 0/1 & & & & TTime & ITime & & & & \\
	\toprule
	\multirow{24}{*}{$\rho > 10^{9}$} & \multirow{8}{*}{\bandone} & CELAR6-SUB0\_20 & 16/57 & 2.00E-03 & 1.00E-03 & 0.17 & \textbf{0.16} & 0.19 & F & F & 97.24 & 0.36 & 20/20 & 4 & 8 & 44/387 \\

	& & CELAR6-SUB1\_20 & 14/75 & 3.00E-04 & 3.00E-04 & \textbf{2.57} & 2.58 & 2.59 & F & F & 444.38 & 0.99 & 20/20 & 5 & 10 & 44/387 \\

	& & CELAR6-SUB2\_20 & 16/89 & 1.00E-04 & 1.00E-04 & 1.05 & \textbf{1.04} & 1.07 & F & F & 653.33 & 0.74 & 20/20 & 5.5 & 11 & 44/387 \\

	& & CELAR6-SUB3\_20 & 18/106 & 1.00E-04 & 1.00E-04 & 3.72 & 3.69 & \textbf{3.67} & F & F & 1219.08 & 0.78 & 20/20 & 5.5 & 11 & 44/387 \\

	& & CELAR6-SUB0\_40 & 16/57 & 3.00E-02 & 2.50E-02 & 4.55 & 4.59 & \textbf{4.17} & F & F & 855.12 & 1.2 & 40/40 & 4 & 8 & 44/774 \\

	& & CELAR6-SUB1\_40 & 14/75 & 1.00E-02 & 1.00E-02 & 392.7 & 388.76 & \textbf{388.42} & F & F & T & T & 40/40 & 5 & 10 & 44/774 \\

	& & CELAR6-SUB2\_40 & 16/89 & 6.50E-03 & 6.40E-03 & 449.02 & 448.56 & \textbf{441.41} & F & F & T & T & 40/40 & 5.5 & 11 & 44/774 \\

	& & CELAR6-SUB3\_40 & 18/106 & 6.00E-03 & 7.00E-03 & 796.76 & 794.98 & \textbf{780.93} & F & F & T & T & 40/40 & 5.5 & 11 & 44/774 \\
	
	\cline{2-17}
	
	& \multirow{11}{*}{\bandtwo} & BN\_30 & 1036/1153 & 15.96 & 5.10E+02 & 1.29 & 1.34 & \textbf{1.03} & F & F & E & E & 50/44.5 & 25 & 41 & 2/4 \\

	& & BN\_31 & 1036/1153 & 47.47 & 2.00E+03 & 1.22 & 1.31 & \textbf{1.05} & F & F & E & E & 50/44.5 & 25 & 39 & 2/4 \\

	& & BN\_32 & 1294/1441 & 1.88 & 1.20E+02 & 2 & 2.01 & \textbf{1.59} & F & F & E & E & 50/44.3 & 28 & 49 & 2/4 \\

	& & BN\_33 & 1294/1441 & 6.20E+02 & 1.00E+03 & 1.93 & 1.96 & \textbf{1.58} & F & F & E & E & 50/44.3 & 26 & 42 & 2/4 \\

	& & BN\_34 & 1294/1443 & 2.60E+04 & 3.20E+04 & 2 & 2.07 & \textbf{1.56} & F & F & E & E & 50/44.3 & 28 & 41 & 2/4 \\

	& & BN\_35 & 1294/1443 & 3.10E+01 & 5.10E+02 & 1.91 & 2.06 & \textbf{1.58} & F & F & E & E & 50/44.3 & 26 & 43 & 2/4 \\

	& & BN\_36 & 1294/1444 & 1.70E+03 & 2.00E+03 & 1.98 & 2.15 & \textbf{1.61} & F & F & E & E & 50/43.7 & 30 & 49 & 2/4 \\

	& & BN\_37 & 1294/1444 & 7.10E+02 & 1.00E+03 & 2.01 & 2.02 & \textbf{1.62} & F & F & E & E & 50/43.7 & 30 & 50 & 2/4 \\

	& & BN\_38 & 1294/1442 & 4.10E+02 & 2.50E+02 & 1.96 & 2.03 & \textbf{1.59} & F & F & E & E & 50/43.9 & 27 & 46 & 2/4 \\

	& & BN\_39 & 1294/1442 & 10.7 & 2.50E+02 & 2.03 & 2.09 & \textbf{1.60} & F & F & E & E & 50/43.9 & 26 & 44 & 2/4 \\

	& & BN\_62 & 657/667 & 7.70E+09 & 8.38E+09 & 0.74 & 0.7 & \textbf{0.68} & F & F & 1.45 & 0.24 & 25/34.1 & 21 & 47 & 2/14 \\
	
	\cline{2-17}
	
	& \multirow{5}{*}{\bandthree} & BN\_60 & 530/539 & 1.20E+08 & 7.20E+16 & \textbf{0.7} & 0.75 & 0.73 & F & F & 1.65 & 0.24 & 50/44.03 & 29 & 60 & 2/16 \\
	
	&  & BN\_61 & 657/667 & 3.00E+10 & 1.70E+10 & 0.74 & 0.74 & \textbf{0.69} & F & F & 1.69 & 0.24 & 25/34.1 & 21 & 46 & 2/14 \\
	
	& & BN\_63 & 530/540 & 1.40E+10 & 9.20E+18 & 2.43 & 0.77 &\textbf{0.67} & F & F & 1.84 & 0.27 & 50/43.5 & 30 & 57 & 2/16 \\

	& & BN\_64 & 530/540 & 1.60E+09 & 5.76E+17 & 0.79 & 0.68 & \textbf{0.63} & F & F & 1.84 & 0.25 & 50/43.5 & 28.5 & 55 & 2/16 \\

	& & BN\_67 & 430/437 & 4.60E+10 & 2.95E+20 & 2.64 & \textbf{1.65} & 2.05 & F & F & 1.71 & 0.26 & 50/50.57 & 32.5 & 62 & 2/16 \\
	\midrule
	\multirow{28}{*}{$\rho \le 10^{9}$} & \multirow{7}{*}{\bandfour} & BN\_20 & 2433/2840 & 119 & 1.00E-04 & 4.53 & \textbf{4.47} & 14.94 & 22.73 & T & T & T & 50/49.3 & 4 & 7 & 91/208 \\

	& & BN\_21 & 2433/2840 & 109 & 1.00E-04 & 4.49 & \textbf{4.42} & 14.86 & 23.37 & T & T & T & 50/49.3 & 4 & 7 & 91/208 \\

	& & BN\_22 & 2119/2423 & 0.97 & 1.00E-05 & \textbf{2.13} & 2.18 & 2.98 & 3.77 & T & 7.83 & 1.71 & 50/47 & 4 & 7 & 91/208 \\

	& & BN\_23 & 2119/2423 & 0.97 & 1.00E-05 & \textbf{2.14} & 2.21 & 3 & 3.74 & T & 7.74 & 1.79 & 50/47 & 2 & 5 & 91/208 \\

	& & BN\_24 & 1514/1818 & 2.08 & 1.00E-05 & \textbf{1.33} & 1.4 & 1.74 & 2.11 & T & 6.24 & 1.58 & 53.8/53 & 2 & 5 & 91/208 \\

	& & BN\_25 & 1514/1818 & 2.01 & 1.00E-05 & \textbf{1.31} & 1.39 & 1.76 & 2.12 & T & 6.34 & 1.62 & 53.8/53 & 2 & 5 & 91/208 \\

	& & Pathfinder & 109/109 & 54.94 & 1.00E-05 & 0.15 & 0.29 & 0.29 & \textbf{0.11} & 0.34 & 0.81 & 0.31 & 52.4/61.4 & 2 & 7 & 63/6437 \\
	
	\cline{2-17}
	
	& \multirow{15}{*}{\bandfive} & Alarm & 37/37 & 3.58 & 11.39 & 0.03 & 0.03 & 0.05 & \textbf{0.02} & 0.06 & 0.46 & 0.21 & 100/99.4 & 2 & 5 & 4/108 \\

	& & Hepar2 & 70/70 & 5.95 & 9 & 0.04 & 0.05 & 0.06 & \textbf{0.03} & 0.19 & 0.42 & 0.19 & 100/100 & 2 & 7 & 4/384 \\

	& & Mildew & 35/35 & 2.00E+03 & 327 & 0.78 & 0.71 & \textbf{0.24} & 0.27 & 2.81 & 2.91 & 1.89 & 75/61.7 & 3 & 5 & 100/14849 \\

	& & Munin & 1041/1041 & 35.77 & 557 & 13.27 & 13.82 & \textbf{1.98} & 3.14 & 11.35 & T & T & 42.1/46.6 & 6 & 9 & 21/276 \\

	& & Munin1 & 186/186 & 43.23 & 16.6 & 598.86 & 629.75 & \textbf{20.6} & 39.01 & A & T & T & 46.2/48.6 & 7 & 12 & 21/276 \\

	& & Munin4 & 1038/1038 & 57.9 & 557 & 16.86 & 17.21 & 2.15 & \textbf{2.06} & 10.67 & 3.77 & 2.01 & 44/46.6 & 6 & 9 & 21/276 \\

	& & Diabetes & 413/413 & 524.51 & 4.24E+06 & 3.28 & 3.31 & \textbf{0.72} & 0.89 & 32.98 & 6.99 & 4.69 & 33.3/45.6 & 4 & 5 & 21/2040 \\

	& & Munin2 & 1003/1003 & 2.90E+05 & 8.90E+08 & 2.71 & 2.28 & \textbf{0.68} & 0.79 & 4.27 & 2.81 & 1.63 & 46.4/48 & 8 & 8 & 21/276 \\

	& & Munin3 & 1041/1041 & 5.00E+05 & 1.20E+04 & 2.71 & 2.52 & \textbf{0.70} & 0.95 & 5.81 & 2.38 & 1.28 & 45.8/37 & 6 & 8 & 21/276 \\

	& & Pigs & 441/441 & 144 & 1.50E+04 & 0.98 & 0.92 & 0.36 & \textbf{0.24} & 1.05 & 1.37 & 0.7 & 55.6/70.2 & 8 & 11 & 3/15 \\

	& & Link & 724/724 & 2.00E+07 & 2.70E+08 & 18.02 & 19.91 & 14.27 & \textbf{3.43} & 29.73 & E & E & 50/65.1 & 12 & 16 & 4/31 \\

	& & Barley & 48/48 & 4.00E+07 & 6.50E+03 & 26.69 & 27.17 & \textbf{1.13} & 1.45 & 15.32 & 17.53 & 10.94 & 100/100 & 4 & 8 & 67/40320 \\

	& & Hailfinder & 56/56 & 13.02 & 1.00E+04 & 0.03 & 0.05 & 0.06 & \textbf{0.01} & 0.05 & 0.53 & 0.23 & 94.2/83.9 & 3 & 5 & 11/1181 \\

	& & Water & 32/32 & 1.00E+05 & 1.00E+06 & 0.17 & \textbf{0.14} & 0.27 & 0.31 & 0.15 & 0.81 & 0.34 & 50/58.23 & 4 & 11 & 4/1454 \\

  & & Win95pts & 76/76 & 8.66 & 3.10E+04 & 0.05 & 0.05 & 0.05 & \textbf{0.03} & \textbf{0.03} & 0.53 & 0.2 & 100/90 & 3 & 9 & 2/252 \\
	\cline{2-17}
	& \multirow{6}{*}{\bandsix} & Andes & 223/223 & 3.10E+04 & 1.50E+20 & 0.57 & 0.59 & 0.19 & \textbf{0.14} & 0.59 & 1.12 & 0.67 & 100/95.7 & 12 & 17 & 2/128 \\

	& & BN\_42 & 870/879 & 1.23 & 4.72E+21 & 32.63 & 32.11 & 35.15 & \textbf{2.66} & 19.18 & 1216.39 & 19.6 & 50/54.4 & 24 & 24 & 2/16 \\

	& & BN\_43 & 870/880 & 1.14 & 3.78E+22 & 65.47 & 64.03 & \textbf{4.37} & 4.43 & A & 1132.1 & 22.24 & 50/54.4 & 25 & 25 & 2/16 \\

	& & BN\_44 & 870/880 & 1.03 & 2.40E+24 & 227.87 & 216.98 & 133.5 & \textbf{12.82} & A & 1341.72 & 17.02 & 50/54.3 & 27 & 27 & 2/16 \\

	& & BN\_45 & 870/880 & 1.1 & 3.78E+22 & 67.05 & 68.21 & 8.07 & \textbf{6.95} & A & 778.91 & 19.05 & 50/54.2 & 25 & 25 & 2/16 \\

	& & BN\_46 & 489/497 & 1.04 & 7.92E+28 & 45.98 & 46.09 & 20.16 & \textbf{5.85} & A & 150.25 & 5.79 & 50/55.9 & 24 & 24 & 2/16 \\

	%& & C432.isc & 432/432 & 4.40E+27 & 7.60E+53 & 45.98 & 46.01 & 34.3 & 31.12 & 105.08 & 141.46 & 2.41 & 50/54.17 & 23 & 28 & 2/512 \\

	%& & C499.isc & 499/499 & 1.04 & 1.27E+30 & 7.52 & 7.49 & 13.25 & 15.77 & A & 11.65 & 0.82 & 50/54.11 & 25 & 25 & 2/32 \\

	%& & C880.isc & 880/880 & 1.16 & 3.78E+22 & 0.53 & 0.53 & 2.43 & 4.91 & A & 1.10 & 0.23 & 50/53.41 & 25 & 25 & 2/16 \\
	\bottomrule
\end{tabular}}
%\end{minipage}
\end{table*}
\end{newEdits}

\begin{newEdits}

\subsubsection{Benchmark experiments}

\paragraph{Benchmark Comparisons.} The results in Table~\ref{table_real_3} are laid out along the lines of Figure~\ref{introTab}. These networks span over a wide range of sparsity (20\% - 100\%), domain sizes (2 -100) and factor arity levels (1-10).

\paragraph{$\rho$ high.} The measure $R_D$ from~\cite{dechter08} predicts superior performance for {\ourModel} only in {\bandone} (CELAR).  However, in this region of high $\rho$, {\ourModel} performs consistently better than the predictions in~\cite{dechter08}. In {\bandone} it is be up to $630$x faster on subsets where ACE completes. In {\bandtwo}, it is up to $2$x faster and in {\bandthree} where the corresponding predictions of~\cite{dechter08} is under-performance by $10^{10}$x - $10^{20}$x,  it can be upto $2.7$x faster than ACE (libDAI and IJGP fail in this space). libDAI fails in these bands since it relies primarily on truth-table indices for its speed which do not scale to these levels of $\rho$.  On the other hand, ACE that takes advantage of {\em factor sparsity} using arithmetic circuits is the only other engine that completes; that said, compiling these structures is costly. We surmise that {\ourModel}'s performance advantages are rooted in the use of multi-way products. Further, applying multi-way products directly on a listing-representation accentuates the gains. 

\paragraph{} The networks in these bands (1, 2 and 3) cover a sparsity range of 20\%-50\% and have factor arity between $1$ to $4$. Further, in Band 1, given that ACE requires the 20\% sparsity levels to complete, we present results at two levels of sparsity for CELAR (20\% and 40\%).  %Also, the networks cover a sparsity range of 20\%-50\% and has factor arity between $1-2$.

%In {\bandtwo} ACE reports runtime failures for BN\_30 - BN\_39.
%In Band $1$, it can be upto $630$x faster while being at least $173$x faster. Further, it is $385$x faster on average. In Band 2, where the corresponding predictions of~\cite{dechter08} is under-performance by $10-$ and 2 it can be upto \yell{1000x} 

\paragraph{$\rho$ low.} $R_D$ predicts superior performance for {\ourModel}  in {\bandfour}, which it achieves. It is upto 5.29x faster than libDAI (it's closest competitor) and up to 5.4x faster than ACE (on the subsets that ACE completes on). IJGP times-out on almost all of the networks. Finally, in Bands 5/6, the two unfavorable settings, {\ourModel} is on an average faster than ACE by $5.8$x/$9.5$x and IJGP by $2.36$x/$2.28$x respectively. It is on average slower than libDAI by $4.8$x/$10$x respectively: libDAI's truth-table indexing advantages clearly manifest in these two bands. We would like to note however, that the corresponding predictions of ~\cite{dechter08} for {\ourModel} is under performance by $10$x - $10^8$x for {\bandfive}, and $10^{20}$x - $10^{28}$x for {\bandsix}, i.e., several orders of magnitude worse.

\paragraph{} Secondly, our fine-grained measure $R_{J}$ (Column 4) is a better predictor for {\ourModel}'s performance than $R_{D}$ (Column 5) on most networks. Consider the Bands 3, 4 and 5 in Table~\ref{table_real_3}. While $R_{D}$ overestimates our performance in \bandfour, $R_{J}$ provides a more realistic measure that correlates with our performance. On the other hand, in Bands 3 and 5, while $R_{D}$ severely underestimates our performance (sometimes by orders of magnitude), $R_{J}$'s predictions are generally tighter. On Bands 1 and 2, its predictions are comparable to $R_{D}$. The only place where it fails to make good estimates is \bandsix.

\paragraph{01-Projections.} %Although $01$ projections are crucial for realizing better bounds in FAQ/AJAR, we empirically demonstrate that they don't present any significant gains in the PGM setting. Although we get non-trivial speedups of $3.2$x on BN\_62 and $\ge 1.5$x on a few other datasets, the average gain is arbitrarily close to 1. 
In theory, 01-projections are central to realizing the asymptotic bounds of FAQ/AJAR. However, we find that in practice they offer mixed results: we found they help in 20/52 networks (average gains of 20\%), they make no difference in 3 networks and marginally hurt in 29 networks (average loss of 7\%). 

\paragraph{Memory Usage.} To examine memory usage, we selected three large networks ($ > 1000$ variables) from Bands 4 and 5 (results in Table~\ref{table_memory}). We found that in the former set {\ourModel} consumed the least memory: on average libDAI consumed 2.35x and ACE consumed 7x more. In the latter set, libDAI consumed the least memory: on average {\ourModel} consumed 2.82x, IJGP 1.98x and ACE 7.29x more. In summary, this underscores that {\ourModel} is comparable to classical inference frameworks on the memory consumption metric.
\begin{table}
	\centering
	\caption{Real-World Experiments: Memory Consumed (in MB)}
	\label{table_memory}
	\begin{tabular}{cccccc}
		\hline \toprule
		Band & Datasets  & \ourModel & IJGP & libDAI  & ACE    \\
		\midrule
		\multirow{3}{*}{\bandthree} & BN\_23 & \textbf{305} & T & 2.35x & 7.41x \\
		& BN\_24 & \textbf{293} & T & 1.67x & 6.83x \\
		& BN\_25 & \textbf{294} & T & 1.71x & 6.95x \\
		\hline
		\multirow{3}{*}{\bandfive} & Munin2 & 395 & 0.56x & \textbf{0.33x} & 3.01x \\
		& Munin3 & 394 & 0.61x & \textbf{0.31x} & 3.05x \\
		& Munin4 & 1038 & 1.03x &\textbf{0.45x}  & 1.37x \\
		\bottomrule
	\end{tabular}%}
	%	\end{minipage}}
\end{table}

%We also examine memory usage for which we selected three large networks each from Bands 4 and 5. 
%To examine memory usage, we select three large networks each from Bands 3 and 5 respectively. We observe that in the former, {\ourModel} consumed the least: on an average, libDAI was over 2x and ACE was 7x. On the latter, libDAI consumed the least: on average, {\ourModel} was 2.7x of libDAI, 1.3x of IJGP; ACE was 2.5x of {\ourModel}. (Table~\ref{table_memory} in supplementary material).
\end{newEdits}

\begin{newEdits}
\paragraph{Hybrid Architecture.} Since {\ourModel} is the only engine that completes on all networks when $\rho$ is high, we now focus on low $\rho$ conditions. As evident from Table~\ref{table_real_3}, HYJAR helps exploit the relative strengths of each strategy--multiway or pairwise products--into a single architecture, yielding consistent performance across a majority of networks (26/28). Of these, in 9 cases (e.g., munin1, munin, barley, mildew) HYJAR's completion times are faster than its nearest standalone competitors ({\ourModel} or libDAI), in 10 cases it is less than 2.5x slower and in 7 cases it is between 2.5x - 4.5x slower. BN\_42 and BN\_44 are the only two networks where HYJAR's strategy does not lead to a notable improvement. Further, follow up analyses indicate that HYJAR consistently switches between strategies at the bag level. We present a detailed analysis of our results in Appendix~\ref{hybrid}. %that multiway was picked more frequently than the pair-wise strategy). %However, in \yell{8} cases HYJAR's completion times are faster than its nearest standalone competitors (e.g. munin1, munin, barley, mildew etc.) indicative of potentially unique advantages accruing from the bagwise heuristic encapsulated in HYJAR. 
% Change Cluster Percentages. Add an algorithm for Pairwise Product. 
% TODO: Get Fraction of Sum of Products as well.
\begin{newEdits}
\begin{table*}[!hbt]
	\centering
	%\resizebox{0.7\textwidth}{!}{
	%\begin{minipage}{0.7\textwidth}
	\caption{{We report \ourModel~w/o 0/1, with all 0/1, HYJAR, Pairwise and LibDAI runtimes (in seconds) along with the total number of clusters, followed by the average \% of clusters running Multiway w/o 0/1, with all 0/1 and Pairwise (i.e., we set $R_{v} = 2$ for all $v \in V$ in Algorithm~\ref{jta}). All runtimes are in seconds.}}
\label{table_hybrid}
	%\resizebox{0.6\textwidth}{!}{
\resizebox{1.0\textwidth}{!}{
\begin{tabular}{ccccccccccc}
	Band & Dataset & \multicolumn{2}{c}{\ourModel} & HYJAR & Pairwise & LibDAI & Clusters & \multicolumn{2}{c}{\ourModel~Clusters (\%)} & Pairwise Clusters (\%)\\
	\hline
	& & w/o 0/1 & 0/1 & & & & & w/o 0/1 & 0/1 & \\
	\toprule
	\multirow{7}{*}{\bandfour} & BN\_20 & 4.52 & 4.53 & 14.94 & 14.47 & 22.73 & 1825 & 0 & 27 & 73 \\

	& BN\_21 & 4.52 & 4.58 & 15.47 & 14.86 & 23.37 & 1825 & 0 & 9 & 91 \\

	& BN\_22 & 2.15 & 2.22 & 2.98 & 2.59 & 3.77 & 1513 & 9 & 32 & 59 \\

	& BN\_23 & 2.16 & 2.24 & 3 & 2.56 & 3.74 & 1513 & 8 & 26 & 66 \\

	& BN\_24 & 1.31 & 1.43 & 1.74 & 1.43 & 2.11 & 908 & 0 & 33 & 67 \\

	& BN\_25 & 1.34 & 1.47 & 1.76 & 1.36 & 2.12 & 908 & 0 & 33 & 67 \\

	& Pathfinder & 0.16 & 0.27 & 0.29 & 0.1 & 0.11 & 91 & 0 & 0 & 100 \\

	\midrule
	\multirow{15}{*}{\bandfive} & Alarm & 0.03 & 0.04 & 0.05 & 0.05 & 0.02 & 27 & 4 & 3 & 93 \\

	& Hepar2 & 0.05 & 0.04 & 0.06 & 0.04 & 0.02 & 58 & 4 & 5 & 91 \\

	& Mildew & 0.81 & 0.76 & 0.24 & 0.21 & 0.27 & 29 & 6 & 6 & 88 \\

	& Munin & 13.34 & 13.61 & 1.98 & 1.74 & 3.14 & 872 & 0 & 6 & 94 \\

	& Munin1 & 600.86 & 630.22 & 20.6 & 19.44 & 39.01 & 158 & 0 & 9 & 91 \\

	& Munin4 & 16.91 & 17.18 & 2.06 & 1.89 & 2.93 & 869 & 0 & 0 & 100 \\

	& Diabetes & 3.31 & 3.36 & 0.72 & 0.6 & 0.89 & 337 & 0 & 7 & 93 \\
	
	& Munin2 & 2.74 & 2.31 & 0.68 & 0.6 & 0.79 & 866 & 0 & 1 & 99 \\
	
	& Munin3 & 2.72 & 2.27 & 0.7 & 0.64 & 0.95 & 901 & 1 & 0 & 99 \\
	
	& Pigs & 1.01 & 0.93 & 0.36 & 0.22 & 0.23 & 368 & 1 & 28 & 71 \\
	
	& Link & 18.02 & 19.91 & 14.27 & 3.28 & 3.44 & 591 & 22 & 12 & 66 \\
	
	& Barley & 26.8 & 27.12 & 1.13 & 0.8 & 1.45 & 36 & 0 & 0 & 100 \\
	
	& Hailfinder & 0.03 & 0.05 & 0.06 & 0.04 & 0.02 & 43 & 0 & 1 & 99 \\
	
	& Water & 0.18 & 0.15 & 0.27 & 0.28 & 0.31 & 19 & 0 & 6 & 94 \\
	
	& Win95pts & 0.04 & 0.03 & 0.05 & 0.06 & 0.03 & 50 & 19 & 13 & 68 \\
	\midrule
	\multirow{6}{*}{\bandsix} & Andes & 0.59 & 0.59 & 0.19 & 0.12 & 0.14 & 178 & 1 & 27 & 73 \\

	& BN\_42 & 32.63 & 33.01 & 35.15 & 2.48 & 2.66 & 789 & 13 & 42 & 45 \\

	& BN\_43 & 65.47 & 64.03 & 4.37 & 4.36 & 4.43 & 789 & 12 & 22 & 66 \\

	& BN\_44 & 227.87 & 216.98 & 133.5 & 16.29 & 12.82 & 789 & 18 & 38 & 44 \\

	& BN\_45 & 67.05 & 68.21 & 8.07 & 8 & 6.95 & 788 & 14 & 25 & 61 \\

	& BN\_46 & 45.98 & 46.01 & 20.16 & 5.64 & 5.85 & 446 & 18 & 17 & 65 \\
	\bottomrule
\end{tabular}}
%\end{minipage}
\end{table*}
\end{newEdits}
%Since {\ourModel} is the only engine (and ACE does on some) that completes when $\rho$ is high, we now focus on low $\rho$ conditions. As evident from Table~\ref{table_real_3}, HYJAR helps exploit the relative strengths of each strategy-multiway or pairwise products- into a single architecture, yielding consistent performance across a majority of networks: 26/28. Of these, in 9 cases (e.g. munin1, munin, barley, mildew etc.) HYJAR's completion times are faster than its nearest standalone competitors \footnote{detailed analysis appears in Table~\ref{table_hybrid} (supplementary  material)}, 10 are less than 2.5x slower and 7 are between 2.5x - 4.5x slower than the best engine ({\ourModel} or libDAI). BN\_42 and BN\_44 are the only two networks where HYJAR's strategy does not lead to a notable improvement. Further, follow up analyses indicate that HYJAR consistently switches between strategies at the bag level (details in Supplementary Material). %that multiway was picked more frequently than the pair-wise strategy). %However, in \yell{8} cases HYJAR's completion times are faster than its nearest standalone competitors (e.g. munin1, munin, barley, mildew etc.) indicative of potentially unique advantages accruing from the bagwise heuristic encapsulated in HYJAR. 

%Update to include consistency aspect.

\paragraph{Takeaways.} (i) In our investigation, we identified a threshold for $\rho$ at $10^9$ reflecting the current memory limits for truth tables (on our machine). While the absolute value of the threshold may change depending on machine configurations, such a threshold will always exist. (ii) In this work,we show that $R_J$ is a better predictor of {\ourModel}'s performance than $R_D$. (iii) Of the 52 networks in the test-bed, HYJAR outperforms libDAI, IJGP and ACE on 39 (i.e., on 75\% of networks). libDAI is the next largest winner with wins on 11 datasets. HYJAR thus offers promise as a practically relevant architecture for building a robust, broadly applicable inference engine.
%HYJAR's outperformance of {\ourModel} and libDAI in 9 networks is indicative of potentially unique advantages accruing from the bagwise heuristic encapsulated in it. HYJAR thus offers promise as a practically relevant architecture for building a robust, broadly applicable inference engine.

\subsubsection{Factor Representations}
As described in Section~\ref{Sec:factorRep}, a secondary representation for our factors is a list of  $\langle \text{index, probability} \rangle$ pairs. Specifically, we use two variants of the list that store indices computed in two variable orders: forward and reverse. We perform three descriptive experiments on the  UAI speech recognition datasets (BN\_20-25) with these new data structures/algorithms, comparing with their corresponding alternatives. Note that the data is stored as compressed indices the memory blowup is only 1.3x (and not 2x) with respect to tries.

\paragraph{Forward vs Reverse Index.} We compare our new method of constructing tries using a reverse index (based on reverse input order) as opposed to the way of constructing tries with a forward index. In particular, this new method removes significant computational/storage overhead and is on an average 3x faster than the older method (see Table~\ref{Rtries}).
\begin{table}
	\caption{Descriptive Experiment: Input Processing. We implement and compare two methods for constructing tries: first using a forward index (Old Tries) and then, using a reverse index (based on reverse input order, New Tries). All runtimes are in seconds.}
	\centering
	\begin{tabular}{cccc}\label{Rtries}
		Dataset & Num Var/D & Old Tries & New Tries \\
		\toprule
		BN\_20 & 2843/91 & 0.1 & 0.27x \\
		BN\_21 & 2843/91 & 0.1 & 0.33x \\
		BN\_22 & 2425/91 & 0.09 & 0.37x \\
		BN\_23 & 2425/91 & 0.09 & 0.38x \\
		BN\_24 & 2425/91& 0.09 & 0.36x \\
		BN\_25 & 2425/91 & 0.09 & 0.36x \\
		\bottomrule
	\end{tabular}
\end{table}

\paragraph{Upward Pass.} During the upward pass (Algorithm~\ref{up}), we store the up message as a hash-table of $\langle$ forward index, reverse index, probability $\rangle$ tuples. Further, we store the cluster tables as lists of $\langle$ reverse index, probability $\rangle$ pairs. Overall, this design gives us on an average 3x speed gains in Algorithm~\ref{up} (see Table~\ref{table:up_pass}) compared to a version that stores both the up messages and cluster tables as lists of $\langle$ forward index, probability $\rangle$ pairs.
\begin{table}
	\centering
	\caption{Descriptive Experiment: Upward Pass. We implement and compare two data structures for storing the up messages/cluster tables: one by storing the up messages as hash-tables of $\langle$ forward index, reverse index, probability $\rangle$ tuples and cluster tables as lists of $\langle$ reverse index, probability $\rangle$ pairs and the other, by storing both up messages/cluster tables as lists of $\langle$ forward index, probability $\rangle$ pairs. All runtimes are in seconds.}
	\begin{tabular}{cccc} \label{table:up_pass}
		Dataset & Num Var/D & New Data Structures & Slowdown of Older Data Structures\\
		\toprule
		BN\_20 & 2843/91 & 1.17 & 7.26x \\
		BN\_21 & 2843/91 & 1.11 & 6.22x \\
		BN\_22 & 2425/91 & 0.27 & 1.33x \\
		BN\_23 & 2425/91 & 0.29 & 1.21x \\
		BN\_24 & 2425/91 & 0.27 & 1.44x \\
		BN\_25 & 2425/91 & 0.28 & 1.46x \\
		\bottomrule
	\end{tabular}
\end{table}

\paragraph{Downward Pass.} Finally, for the downward pass (Algorithm~\ref{down}), we perform an in-place $\hash$ and compare it with a version implementing Sort-MergeProduct. In particular, we observe that using $\hash$s instead of Sort-MergeProduct gives us an average speedup of 1.6x (see Table~\ref{table:down_pass}).
\begin{table}
	\centering
	\caption{Descriptive Experiment: Downward Pass. We implement and compare two algorithms for computing products in the downward pass: $\hash$ and SortMergeProduct.  All runtimes are in seconds.}
	\begin{tabular}{cccc} \label{table:down_pass}
		Dataset & Num Var/D & $\hash$ & Slowdown of Sort-MergeProduct \\
		\toprule
		BN\_20 & 2843/91 & 0.53 & 3.02x \\
		BN\_21 & 2843/91 & 0.52 & 2.58x \\
		BN\_22 & 2843/91 & 0.11 & 1.18x \\
		BN\_23 & 2843/91 & 0.12 & 1.17x \\
		BN\_24 & 2843/91 & 0.17 & 1.06x \\
		BN\_25 & 2843/91 & 0.18 & 1.06x \\
		\bottomrule
	\end{tabular}
\end{table}

\begin{newEdits}
\section{Conclusion and Future Directions}
%In this paper we have shown that recent theoretical improvements in exact inference based on new database join algorithms lead to speedup in %exact inference for 
%PGMs with large domain sizes and sparse factors. The following future research directions seem promising: (1) %Come up with a heuristic that allows one to combine 
%Develop heuristics that enable combining \ourModel\ with other engines (e.g. ACE, IJGP and libDAI) %that work better with
%better suited to small domain sizes and less sparse factors; (2) Incorporate Single Instruction Multiple Data (SIMD) instructions to speed up the computation in a bag (this has been successful in database joins~\cite{eh}); and finally (3) utilize \ourModel\ to speed up approximate inference engines (though 01-projections can be extended to approximate inference, there are currently no known approximate extensions for the multi-way product algorithm).

%In this paper we show that GHD based inference algorithms do show promise in practical settings. Specifically, we operationalize recent theoretical improvements in exact inference based on new database join algorithms in a prototype 

This paper demonstrates that in a wide range of PGM benchmarks, GHD based inference algorithms offer much promise in terms of performance and prior conclusions on their practical (ir)relevance~\cite{dechter08} might have to be re-visited. Further, the HYJAR architecture shows great promise in integrating the benefits of the traditional inference engines along with {\ourModel}. Improving the data structures of {\ourModel} to facilitate this integration is one of our future work. The following are other future research directions that seem promising: (1) Incorporate Single Instruction Multiple Data (SIMD) instructions to speed up the computation in a bag (this has been successful in database joins~\cite{eh}); and (2) utilize \ourModel\ to speed up approximate inference engines (there are currently no known approximate extensions for the multi-way product algorithm).

\end{newEdits}

\subsection*{Acknowledgments}
We thank Andrew Lamb, Chris Aberger, Hung Ngo, Jimmy Dobler and Ravishankar Krishnaswamy for helpful discussions. 

\paragraph{} SVMJ's research is supported in part by NSF grant CCF-1717134 and thanks Microsoft Research for hospitality where a part of this work was done. CR gratefully acknowledges the support 
of DARPA No. N6600115C4043 (SIMPLEX), No. FA87501720095 (D3M), No. FA87501220335 (XDATA), No. FA87501320039 (DEFT), DOE under No. 108845 (Integrated Compiler and Runtime Autotuning Infrastructure for Power, Energy, and Resilience), NSF under No.1505728 (Intel/NSF CPS Security grant), NIH under No. U54EB020405 (Mobilize), ONR under No. N000141712266 (Unifying Weak Supervision), No. N00014-14-1-0102 (Data-Driven Systems: Join Algorithms and Random Network Theory), AFOSR under No. 580K753 (Mathematical Foundations of Secure Computing Clouds), Moore Foundation, Okawa Research Grant, American Family Insurance, Accenture, Toshiba, Secure Internet of Things Project, Google, VMware, Qualcomm, Ericsson, Analog Devices, and members of the Stanford DAWN project: Intel, Microsoft, Teradata, and VMware. The U.S. Government is authorized to reproduce and distribute reprints for Governmental purposes notwithstanding any copyright notation thereon. Any opinions, findings, and conclusions or recommendations expressed in this material are those of the authors and do not necessarily reflect the views, policies, or endorsements, either expressed or implied, of DARPA, DOE, NSF, NIH, ONR, or the U.S. Government.
%of Defense Advanced Research Projects Agency (DARPA) under agreement number FA8750-17-2-0095, DARPA SIMPLEX program under No. N66001-15-C-4043, DARPA FA8750-12-2-0335 and FA8750-13-2-0039, DOE 108845, National Institute of Health (NIH) U54EB020405, the National Science Foundation (NSF) under award No. CCF-1563078, the Office of Naval Research (ONR) under awards No. N00014-17-1-2266, the Moore
%Foundation, the Okawa Research Grant, American Family Insurance, Accenture, Toshiba, and Intel. Any opinions, findings, and conclusions or recommendations expressed in this material are those of the authors and do not necessarily reflect the views
%of DARPA, NSF, ONR, or the U.S. government. 
AR's research is supported in part by NSF grants CCF-1763481 and CCF-1717134.

%\bibliographystyle{acm}
%\bibliography{atri_short}

%{\footnotesize
\bibliographystyle{abbrv}
\bibliography{atri_short}
%}

\newpage
\appendix
%\onecolumn

\section{Missing Details in Section 3}

\subsection{01-Projections}  \label{01projections}
As stated earlier, using 01-Projections is central to realizing the asymptotically better bounds in FAQ/AJAR. Moreover, this gain is accentuated when we deal with sparsity: the key idea is to exploit the \emph{sparsity of all the input factors} and not just those encompassed by the query corresponding to the current bag. We highlight this aspect in an example where the gain is more dramatic than Example \ref{eg:bag2}.
\begin{example}
	\label{ex:appendix}
	For instance, say we are processing the bag of a GHD $T$ at node $v$ that contains $k$ factors $\mathcal{K}_v = \{\phi_1(X_{1}, X_{2}),\phi_2(X_{1}, X_{3}), ..., \phi_k(X_{1}, X_{k+1})\}$ and $k+1$ variables $\chi(v) = \{X_{1}, X_{2},...,X_{k+1}\}$, and, that the query for the bag involves marginalizing out variable $X_{1}$ to create an intermediary factor/message $\phi\rq{}(X_{2},X_{3},...,X_{k+1})$. The optimal fractional cover (recall the LP in~\eqref{eq:lpAGM}) for the query hypergraph would be $1$ for all $k$ edges making the runtime complexity of processing this query $O(N^k)$, a bound that could be significant depending on the value of $k$.
	
	Now suppose the original PGM hypergraph $\mathcal{H}$ contains a factor %on a subset of the variables in $\chi(v)$, 
	$\psi(X_{2}, X_{3},..., X_{k+1}) \not\in \lambda(v)$ that would be encountered somewhere along the tree $T$ in subsequent computations. We could employ this factor upfront while computing query at $v$ to avoid redundant computations at a later stage. With the addition of this factor support the optimal fractional cover of the induced query hypergraph would be $1-\frac{1}{k}$ for the new hyperedge $\{X_{2}, X_{3},...,X_{k+1}\}$ and $\frac{1}{k}$ for the other $k$ edges incident on $X_1$. %Thus, the runtime bound of processing this query is now $O(N^{2 - \frac{1}{k}})$, a substantial speed-up especially if $N$ and $k$ are large.
	This makes the AGM bound go to $O(N^{2 - \frac{1}{k}})$, a dramatic improvement in the runtime bound especially if $N$ and $k$ are large.
	
	However, employing this factor as-is could lead to double counting of probabilities while processing the additional factor. Hence, we need to find a way of incorporating such support factors without changing the computation. % and bounds of the original query.
\end{example} 

We accomplish this goal using 01-projections defined by:
%
%In general, 01-projections are defined by:
\begin{definition}
	For each $S \in \mathcal{E}$ and any set $U \subseteq \mathcal{V}$ such that $S \cap U \neq \emptyset$, define the 01-projection of $\phi_S$ onto $U$ as the function
	\begin{equation*}
	\phi_{S/U}\text{:} \prod\limits_{U\rq{} \in S \cap U} D(U\rq{}) \rightarrow \{0,1\},
	\end{equation*}
	where
	\begin{equation*}
	\phi_{S/U}(\boldsymbol{\mathrm{x}}_{S \cap U}) = \begin{cases} 1 \text{ if } \text{there exists } \boldsymbol{y}_S \text{ s.t. } \phi_S(\boldsymbol{y}_S) \neq 0 \text{ and } \boldsymbol{y}_{S \cap U} = \boldsymbol{\mathrm{x}}_{S \cap U} \\
	0 \text{ otherwise } \end{cases}
	\end{equation*}
	
\end{definition}
%\begin{newEdits}
Two key improvement afforded by $01$-projections can be summarized as follows:
\begin{itemize}
	\item \emph{In terms of computation}: there could be potential wastage in computing entries that would eventually be annihilated. We avoid this redundancy by computing only those entries that we know will not be eliminated. When dealing with very large factors exploiting \emph{factor sparsity} via projections could lend substantial reductions in computation.
	%When dealing with very large, sparse factors this could lend substantial reductions in computation.
	
	\item The above also implies \emph{better theoretical bounds}: by incorporating support factors using 01-projections while processing the query corresponding to each bag, we are now bounded by the fractional cover of the induced hypergraph formed by using the projection of edges incident on $\chi(v)$ in addition to edges in $\lambda(v)$. This gain as illustrated in Example~\ref{ex:appendix} can be substantial.
\end{itemize}

\subsection{GHD: Notions of Width} \label{ghdnotionsSupp}

%A tree-decomposition of hypergraph $\mathcal{H}$ is defined by a tuple $ \langle T,\chi \rangle$, where $T(V(T), E(T))$ is a tree, $\chi: V(T) \rightarrow 2^{\mathcal{V}}$ is a function associating a set of vertices $\chi(v) \subseteq \mathcal{H}$ to each node $v$ of $T$ such that the following properties hold (i) for each $e \in \mathcal{E}$, there is a node $v \in V(T)$ such that $e \subseteq \chi(v)$, and, (ii) for each $t \in \mathcal{V}$, the set $\{v \in V(T)|t \in \chi(v)\}$ is connected in $T$.

%The \emph{tree-width} of a tree decomposition $\langle T,\chi \rangle$ of $\mathcal{H}$ is $max\{|\chi(v)| - 1 | v \in V(T)\}$, i.e., it is the maximum bag-size of the decomposed tree \emph{minus} 1. The \emph{tree-width} of $\mathcal{H}$, denoted by $tw(\mathcal{H})$, is the minimum of the tree-widths of all tree decompositions of $\mathcal{H}$.

We describe three notions of width here -- treewidth ($\tw$), hypetreewidth ($\htw$) and fractional hypertreewidth ($\fhtw$) for a GHD $G = (T, \chi, \lambda)$ (recall Definition~\ref{Definition:ghd}).  The \emph{treewidth} of a GHD is given by \\ $\max_{v \in V(T)}(|\chi{(v)}|)$. \footnote{The standard definition of \emph{treewidth} is $\max_{v \in V(T)}(|\chi{(v)}|) - 1$ but we use our modified definition thoroughout the paper.} The notions $\htw$ and $\fhtw$ used in this paper can be derived from the concept of \emph{edge covers} of a hypergraph~\cite{faq,agm}. To this end, for a hypergraph $\mathcal{H} = (\mathcal{V},\mathcal{E})$, let $B \subseteq \mathcal{V}$ be any subset of vertices. We define $\gamma_{\mathcal{H}}(B)$ and $\gamma^*_{\mathcal{H}}(B)$ to be the minimum \emph{integral} edge cover and the minimum \emph{fractional} edge cover respectively of $B$ using edges in $\mathcal{E}$. In particular, $\gamma^*_{\mathcal{H}}(B)$ is the optimal objective value of LP~\eqref{eq:lpAGM} obtained by replacing $|\phi_{S}|$ with $\max_{S \in \mathcal{E}}|\phi_{S}|$ for every $S \in \mathcal{E}$. Finally, $\gamma_{\mathcal{H}}(B)$  is the optimal objective value of the following integer program:
\begin{eqnarray*} 
	\text{min}  && \sum_{S \in \mathcal{E}} x_S \\
	\text{s.t.} && \sum_{S:v \in S} x_S \geq 1, \forall v \in B \\
	\text{} && x_S \in \{0,1\}, \forall S \in \mathcal{E}.
\end{eqnarray*} 
Then, we have $\htw = \max_{v \in V}\gamma_{\mathcal{H}}(\chi(v))$ and $\fhtw = \max_{v \in V}\gamma^*_{\mathcal{H}}(\chi(v))$. Note that the following inequality is always true for a fixed $G$:
\begin{equation} \label{eq:fhtwinequality}
\fhtw \le \htw \le \tw.
\end{equation}

\section{Missing Details in Section 4}

\subsection{Comparison Engines: Compile and Run Commands} \label{compilerunc}
 We outline the commands that we use to run the various inference engines.

			To run IJGP, we run \texttt{./ijgp --i-bound <No\_of\_Variables> --order min-fill <InputFilename.uai> <EmptyEvidenceFile>}.
			
			To run LibDAI, we run \texttt{./example <InferenceFlag=0> <InputFilename.fg> <OutputFilewith\\Marginals>}.
			
			For ACE with IJCAI-2005 Datasets in the BN Learn repository (Munin1-4 and diabete)s, we run \texttt{compile -cd05 -dtClauseMinfill <InputFilename.uai>}, followed by \texttt{evaluate <InputFilename.uai>}, setting a memory limit of $54$GB (which is $85\%$ of our total RAM capacity as recommended in~\cite{aceweb}).
			
			For ACE with other IJCAI-2005 Datasets in the BN Learn repository than the above, we run \texttt{compile -cd05 -dtBnMinfill <InputFilename.uai>}, followed by \texttt{evaluate <InputFilename.uai>}, setting a memory limit of $54$GB (which is $85\%$ of our total RAM capacity as recommended in~\cite{aceweb}).
			
			For ACE with all other datasets, we run \texttt{compile <InputFilename.uai>}, followed by \texttt{evaluate <Input\\Filename.uai>}. Whenever we run into a memory error, we set it to $54$GB (\cite{aceweb}) and rerun our code (if it runs into the same error in this run, we stop).

\subsection{Detailed Analysis of Table~\ref{table_hybrid}} \label{hybrid}
From Table~\ref{table_hybrid}, we first observe that HYJAR picks mixtures of the three choices ({\ourModel} (with/without 01) and {\PP}) on most benchmarks i.e., it frequently switches between the strategies on most networks. When our Algorithm~\ref{hybridAlgo} leans more consistently towards the better strategy, HYJAR demonstrates good gains -- for example, consider the Munin subsets 1-4, Barley and Diabetes. In these networks, our heuristic actually picks {\PP} for 95\% of the bags on average. On the other hand, when the algorithm leans less strongly towards the better strategy, for instance, in datasets like BN\_20/21 and BN\_42,45, HYJAR does not make significant gains. In some cases, the cost of switching between data structures needed for {\NPRR} and {\PP} is not compensated for by the benefits of the hybrid system. Further, our numbers reveal very interesting insights -- for example, consider the networks BN\_43 and BN\_45. In both of them multiway products are much slower compared to pairwise, but the overall HYJAR numbers are extremely competitive (in spite of running multiway products on more than a third of the clusters). Overall, we believe that HYJAR is promising in the direction of building adaptive inference engines that are optimal for a given input PGM, automatically tuning themselves to its characteristics.
\end{newEdits}

\end{document}